\documentclass{article}

\usepackage[main, final]{neurips_2026}

\usepackage[utf8]{inputenc} 
\usepackage[T1]{fontenc}    
\usepackage{hyperref}       
\usepackage{url}            
\usepackage{booktabs}       
\usepackage{amsfonts}       
\usepackage{nicefrac}       
\usepackage{microtype}      
\usepackage{xcolor}         

\usepackage{multirow}
\usepackage{array}
\usepackage{makecell}
\usepackage{amsmath}
\usepackage{cleveref}
\usepackage{graphicx}
\usepackage{subcaption}
\usepackage{wrapfig}

\title{The Dormant Spiking Neuron: A State-Driven Mechanism for Efficient Spiking Neural Networks}

\author{%
  Ruichen Ma$^1$, Yicai Chen$^1$, Pujun Zhou$^1$, Yue Zuo$^1$, Chongyang Li$^1$, Shuang Liu$^1$,\\
  \textbf{Shaogang Hu$^{1,2}$, Guanchao Qiao$^1$\thanks{Corresponding author.}} \\
  $^1$University of Electronic Science and Technology of China (UESTC)\\
  $^2$Shenzhen Institute for Advanced Study, UESTC\\
  \texttt{ruichen.ma@std.uestc.edu.cn, gcqiao@uestc.edu.cn} \\
  {\hypersetup{pdfborder={0 0 0}}\href{https://github.com/Ruichen0424/DSN}{\textcolor[HTML]{0066CC}{https://github.com/Ruichen0424/DSN}}}
}

\begin{document}

\maketitle

\begin{abstract}

Spiking neural networks (SNNs) offer a compelling pathway toward energy-efficient artificial intelligence, fundamentally anchored in presynaptic input-driven sparsity.
However, a substantial yet chronically overlooked source of computational overhead persists: the continuous subthreshold dynamics of postsynaptic neurons.
Quantitative analysis reveals that up to 44\% of neurons in an ImageNet-trained ResNet-18 model reside in a deeply inhibited state, imposing a persistent computational burden despite having a negligible probability of firing.
To address this inefficiency, the Dormant Spiking Neuron (DSN) model is introduced.
By incorporating a state-driven dynamic gate, the DSN allows a neuron to enter a dormant state when its membrane potential falls below a predefined negative threshold, dynamically pruning redundant computations at the instance level.
Crucially, this dormancy mechanism is directly integrated into the neuronal dynamics and trained end-to-end via backpropagation through time using surrogate gradients, enabling the network to actively learn a robust dormancy policy that adaptively preserves critical information flow.
Extensive experiments across static and neuromorphic datasets demonstrate that the DSN framework consistently reduces computational energy by 40\% to over 70\%.
Notably, on the large-scale ImageNet dataset, the DSN outperforms state-of-the-art static pruning techniques at comparable reported cost reductions and simultaneously improves the corresponding baseline SNN accuracy by 1.34\%.
By establishing a trainable, state-driven sparsity paradigm orthogonal to conventional input-driven methods, the DSN unlocks synergistic efficiency gains, presenting an effective approach for advancing neuromorphic computing.

\end{abstract}

\section{Introduction}
\label{sec:introduction}

The escalating demand for artificial intelligence on resource-constrained platforms \cite{atzori2010internet, chen2019deep, shuvo2022efficient, aouedi2024survey, ma2024b} has catalyzed significant research into energy-efficient computing paradigms.
Among these, spiking neural networks (SNNs) have emerged as a compelling alternative to conventional artificial neural networks (ANNs), owing to their biologically plausible, event-driven computational nature \cite{maass1997networks, pfeiffer2018deep, xu2018csnn, roy2019towards, tavanaei2019deep, zenke2021visualizing, deng2024spiking, ma2026i2e, ma2026intrinsically}.
Fundamentally, the efficiency of SNNs is anchored in input-driven sparsity, a presynaptic mechanism wherein computations are triggered exclusively by the arrival of discrete spike events.
This mechanism effectively bypasses operations associated with inactive inputs, leading to substantial reductions in the total number of synaptic operations compared to the dense matrix multiplications characteristic of ANNs \cite{merolla2014million, furber2014spinnaker, qiao2015reconfigurable, davies2018loihi, imam2020rapid}.

Despite this advantage, a substantial yet chronically overlooked source of computational overhead persists within the SNN framework, stemming from the continuous subthreshold dynamics of postsynaptic neurons.
Even when a receiving neuron remains quiescent, it still performs membrane-potential integration and leakage calculations at every timestep \cite{gerstner2002spiking}.
To quantify the magnitude of this inefficiency, we analyzed the membrane-potential distribution in a ResNet-18 SNN model trained on the ImageNet dataset.
The cumulative distribution function of neuronal membrane potentials reveals three distinct regions (see the Appendix \ref{app_c1} and \cref{fig:app_1} for a detailed visualization).
While approximately 26\% of neurons are in the active spiking region and 30\% reside in the subthreshold active region ($0 < V_m < V_{th}$), which is critical for potential future spikes, 44\% of the neurons are found in the deeply inhibited region ($V_m \le 0$).
These deeply inhibited neurons have a negligible probability of firing in the immediate future.
Consequently, the computational resources consistently dedicated to updating their states represent an opportunity for optimization.

Targeting this specific source of inefficiency, this paper introduces the Dormant Spiking Neuron (DSN) model.
By incorporating a state-driven dynamic gate, the DSN allows a neuron to enter a dormant state when its membrane potential falls below a predefined negative threshold, dynamically excluding it from all subsequent computations within that sample.
This postsynaptic sparsity operates orthogonally to conventional presynaptic input-driven sparsity, enabling a multiplicative reduction in overall computational load.
The DSN differs from conventional static pruning techniques.
While traditional network pruning aims to remove structures based on dataset-level, chronic redundancy \cite{shi2024towards, li2024towards}, the DSN leverages state-driven sparsity to target instance-level, transient redundancy.
Recent advances in neuromorphic hardware have demonstrated the potential for physical energy savings from threshold-based neuron inhibition mechanisms \cite{zhou2025energy}.

However, static, post-hoc thresholding inherently assumes that early inhibition implies future unimportance.
Although this approach can be effective for stationary inputs, it may be less robust and can lead to accuracy degradation in complex or nonstationary temporal streams.
To address this limitation, the dormancy mechanism of the DSN is directly integrated into the neuronal dynamics and trained end-to-end via backpropagation through time (BPTT) using surrogate gradients (SG).
This allows the network to adapt its weight distribution to the dormancy constraint and learn a robust dormancy policy that preserves critical information flow.
The primary contributions of this work are summarized as follows:
\begin{itemize}
\item The DSN is proposed as a state-driven sparsity paradigm that specifically targets the computational inefficiency of deeply inhibited neurons. By introducing dynamic, instance-level neuronal dormancy, the DSN operates orthogonally to conventional presynaptic input-driven sparsity, thereby enabling a multiplicative reduction in overall synaptic operations.
\item The BPTT training methodology for the DSN is formulated using surrogate gradients, thereby mitigating the performance degradation that can arise from static or post-hoc thresholding and allowing the network to learn a robust dormancy policy.
\item Extensive experiments on multiple static and neuromorphic datasets demonstrate that the DSN reduces computational energy by 40\% to over 70\% while maintaining competitive performance. Notably, on the ImageNet dataset, the proposed method outperforms state-of-the-art static pruning techniques at comparable reported cost reductions while simultaneously improving the corresponding baseline SNN accuracy by 1.34\%.
\end{itemize}

\section{Related Work}
\label{sec:related_work}

Various pruning techniques aim to improve SNN efficiency by identifying and removing less critical network components based on static properties or observed activity \cite{chen2021pruning, chowdhury2022towards, chakraborty2024sparse, shi2024towards, li2024towards, wei2025qp}.
Common approaches include unstructured pruning, such as eliminating synapses with small weight magnitudes, and structured pruning, which removes entire neurons based on criteria such as output contribution.
Within the SNN domain, these methods often use neuronal firing rates, permanently eliminating components that fall below a specified firing-rate threshold.
While effective at reducing overall network size, these conventional pruning strategies predominantly target dataset-level, chronic redundancy.
By relying on statistical measures of importance derived from representative datasets, these methods have limited adaptability to dynamic inputs, where the relevance of a specific neuron may vary rapidly across samples.
Furthermore, decisions based solely on long-term firing rates do not capture the real-time computational overhead associated with the subthreshold state updates of temporarily inhibited neurons, leaving a substantial portion of computational redundancy unaddressed.

The proposed DSN differs from conventional static pruning by introducing dynamic, state-driven sparsity.
Instead of permanently removing structural components, the DSN dynamically bypasses redundant computations when a neuron's membrane potential falls below a predefined negative threshold ($V_m < D_{th}$).
This mechanism specifically targets instance-level, transient redundancy, offering flexibility and adaptability compared with statistical pruning methods.
Because the DSN regulates postsynaptic computations based on the internal state of the receiving neuron, it operates orthogonally to conventional presynaptic input-driven sparsity.
This orthogonality enables the two forms of sparsity to act synergistically, yielding multiplicative reductions in computational cost.

\begin{figure}[!t]
    \centering
    \includegraphics[width=\textwidth]{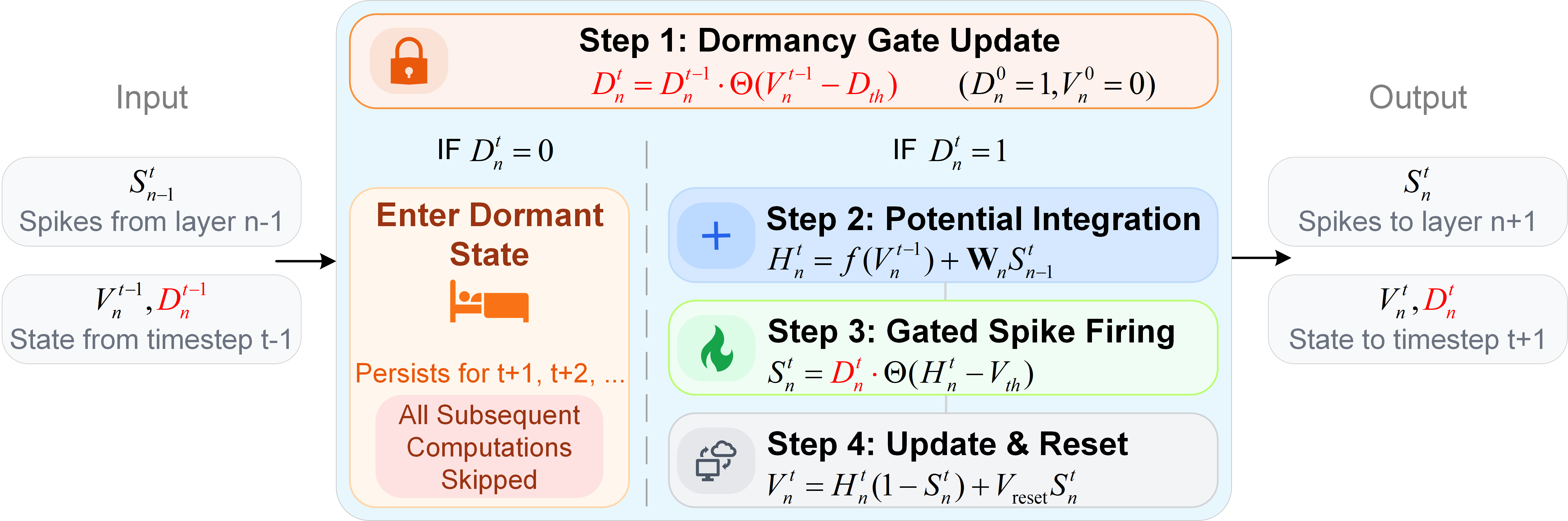}
    \caption{
      The four-step dynamics of the Dormant Spiking Neuron (DSN).
      At each timestep $t$, a DSN executes a sequence of operations.
      (1) Dormancy Gate Update: The neuron first checks whether its previous membrane potential $V_n^{t-1}$ has fallen below the dormancy threshold $D_{th}$.
      If so, it enters a permanent dormant state, and all subsequent stages are bypassed for the current and all future timesteps.
      (2) Potential Integration: If active ($D_n^t=1$), the neuron integrates presynaptic inputs with its decayed membrane potential.
      (3) Gated Spike Firing: A spike $S_n^t$ is generated only if the neuron is active and its integrated potential $H_n^t$ surpasses the firing threshold $V_{th}$.
      (4) Potential Update \& Reset: The membrane potential $V_n^t$ is updated for the next timestep, incorporating a reset mechanism.}
    \label{fig:dsn_dynamics}
\end{figure}

\section{Method}
\label{sec:method}

\subsection{Dormant Spiking Neuron Dynamics}
\label{sec:dsn_model}
The core of the DSN involves introducing a binary activity state, termed the dormancy gate $D_n^t \in \{0, 1\}$, for each neuron in layer $n$ at timestep $t$.
This gate dynamically determines whether a neuron should enter a permanent state of inactivity, thereby bypassing its associated computations.
The complete forward dynamics of a DSN follow a four-stage process at each timestep, as shown in \cref{fig:dsn_dynamics}.

\textbf{Step 1: Dormancy Gate Update.}
The activity state $D_n^t$ is updated based on the membrane potential from the previous timestep $V_n^{t-1}$.
If $V_n^{t-1}$ drops below a predefined dormancy threshold $D_{th}$, the gate irreversibly closes ($D_n^t \to 0$).
This update is described by:
\begin{equation}
  D_n^t = D_n^{t-1} \cdot \Theta(V_n^{t-1} - D_{th})
  \label{eq:dormancy_update}
\end{equation}
Initially, all neurons start in an active state ($D_n^0=1$) with their membrane potential at rest ($V_n^0=0$).
The multiplicative nature of \cref{eq:dormancy_update} ensures that once the gate closes ($D_n^t=0$), it remains permanently closed for all subsequent timesteps ($D_n^{t_0} \equiv 0$ for any $t_0>t$), rendering the neuron dormant.

\textbf{Step 2: Potential Integration.}
Active neurons ($D_n^t=1$) integrate weighted spikes $S_{n-1}^t$ from the presynaptic layer with their decayed membrane potential $f(V_n^{t-1})$ from the previous timestep:
\begin{equation}
	H_n^t = f(V_n^{t-1}) + \mathbf{W}_n S_{n-1}^t
	\label{eq:potential_integration}
\end{equation}

\textbf{Step 3: Gated Spike Firing.}
A spike $S_n^t$ is generated only if the neuron is active ($D_n^t = 1$) and its integrated potential $H_n^t$ surpasses the firing threshold $V_{th}$:
\begin{equation}
	S_n^t = D_n^t \cdot \Theta(H_n^t - V_{th})
	\label{eq:gated_firing}
\end{equation}

\textbf{Step 4: Potential Update \& Reset.}
The membrane potential $V_n^t$ for the next timestep is calculated, incorporating a reset mechanism for neurons that have fired:
\begin{equation}
	V_n^t = H_n^t (1 - S_n^t) + V_{\text{reset}} S_n^t
	\label{eq:potential_reset}
\end{equation}
Here, $\Theta(\cdot)$ is the Heaviside step function, $\mathbf{W}_n$ is the weight matrix from layer $n-1$ to $n$, and $f(\cdot)$ denotes the membrane potential decay function.
For a standard Leaky Integrate-and-Fire (LIF) neuron model, $f(V)$ is typically defined as $f(V) = \tau^{-1}\cdot V$, where $\tau$ is the membrane time constant.

Compared to a standard spiking neuron, the DSN incorporates two primary modifications.
These are the dormancy gate update mechanism specified in \cref{eq:dormancy_update} and the gating of the spike output by $D_n^t$ in \cref{eq:gated_firing}.
A crucial consequence of this design is that a dormant neuron will never fire again.
Therefore, subsequent dynamics for that neuron, including membrane-potential updates, need not be executed.
This property underlies the computational savings provided by the DSN.

\subsection{Direct Training via Surrogate Gradients}
\label{sec:training}

A primary challenge in training SNNs arises from the non-differentiable nature of the Heaviside step function $\Theta(\cdot)$, which appears in both \cref{eq:dormancy_update,eq:gated_firing}.
The surrogate-gradient method enables gradient-based training despite the non-differentiability of the Heaviside step function $\Theta(\cdot)$ \cite{neftci2019surrogate}.
The core principle involves using the exact function $\Theta(\cdot)$ in the forward pass but substituting its derivative with a continuous approximation, the surrogate function $\sigma'(\cdot)$, during the backward pass.

This enables the derivation of a complete set of update rules for backpropagation through time.
The primary goal of BPTT is to compute the gradient of the loss function $L$ with respect to the weights $\mathbf{W}_n$, which is achieved by summing the contributions from each timestep:
\begin{equation}
\frac{\partial L}{\partial \mathbf{W}_n} = \sum_{t=1}^{T} \frac{\partial L}{\partial H_n^t} \frac{\partial H_n^t}{\partial \mathbf{W}_n} = \sum_{t=1}^{T} \delta_{H,n}^t (S_{n-1}^t)^\top
\label{eq:weight_gradient}
\end{equation}
The weight gradient depends on the error term $\delta_{H,n}^t \equiv \frac{\partial L}{\partial H_n^t}$, which represents the gradient of the loss concerning the integrated potential $H_n^t$.
The term $\delta_{H,n}^t$ is part of a system of interconnected gradients that must be determined by propagating errors backward from the network output.
To formalize this process, error terms are defined for the other key variables, including $\delta_{S,n}^t$ for the spike output $S_n^t$, $\delta_{V,n}^t$ for the updated potential $V_n^t$, and $\delta_{D,n}^t$ for the dormancy gate state $D_n^t$.
These errors propagate backward through spatial layers and temporal steps according to the following recurrence relations:
\begin{align}
\left\{
\begin{aligned}
	\delta_{H,n}^t &= \delta_{S,n}^t \cdot [D_n^t \cdot \sigma_s'(H_n^t - V_{th})] + \delta_{V,n}^t \cdot (1 - S_n^t) \\
	\delta_{S,n}^t &= \mathbf{W}_{n+1}^\top \delta_{H, n+1}^t \\
	\delta_{V,n}^t &= \delta_{D,n}^{t+1} \cdot [D_n^t \cdot \sigma_d'(V_n^t - D_{th})] + \delta_{H,n}^{t+1} \cdot f'(V_n^t) \\
	\delta_{D,n}^t &= \delta_{D,n}^{t+1} \cdot \Theta(V_n^t - D_{th}) + \delta_{S,n}^t \cdot \Theta(H_n^t - V_{th})
\end{aligned}
\right.
\end{align}
Here, $\sigma_s'(\cdot)$ and $\sigma_d'(\cdot)$ represent the surrogate gradient functions used for spike firing and dormancy gate mechanisms, respectively.
The backpropagation process begins at the output layer at the final timestep and iteratively computes gradients that are propagated backward through time and space.
Solving this system yields the required $\delta_{H,n}^t$ terms at each timestep, which are then used to compute the final weight gradient via \cref{eq:weight_gradient}.
Full mathematical derivations are available in the Appendix \ref{app:bptt_derivation}.

\begin{figure}[!t]
  \centering
  \begin{subfigure}{0.245\linewidth}
    \centering
    \includegraphics[width=1\textwidth]{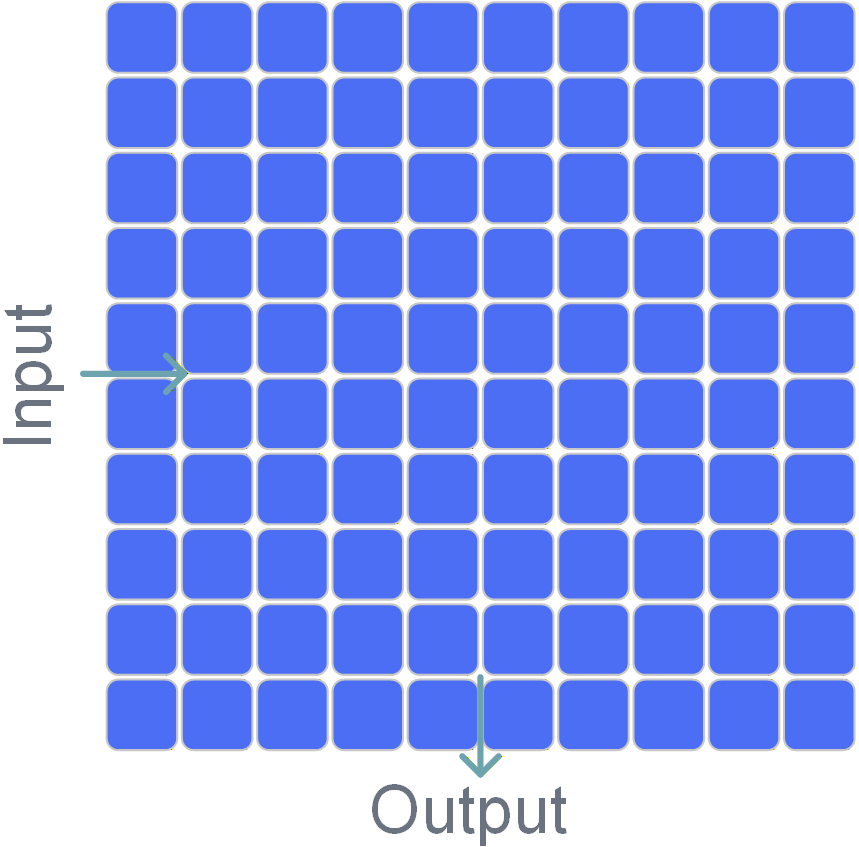}
    \caption{Dense operation.}
    \label{fig:3a}
  \end{subfigure}
  \centering
  \begin{subfigure}{0.245\linewidth}
    \centering
    \includegraphics[width=1\textwidth]{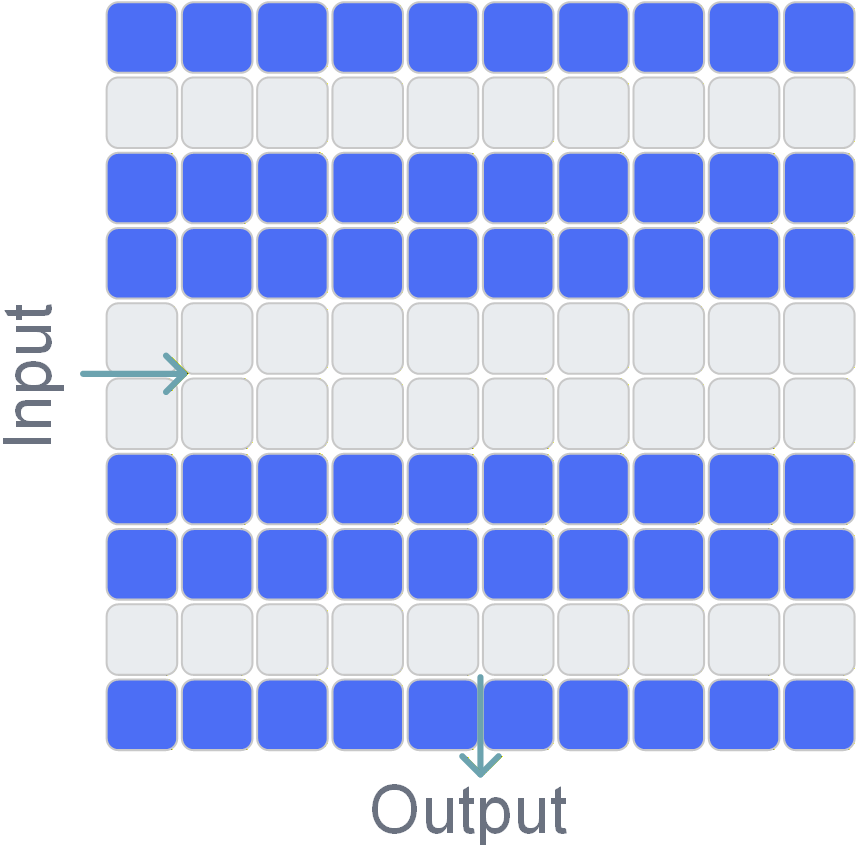}
    \caption{Input-driven sparsity.}
    \label{fig:3b}
  \end{subfigure}
  \begin{subfigure}{0.245\linewidth}
    \centering
    \includegraphics[width=1\textwidth]{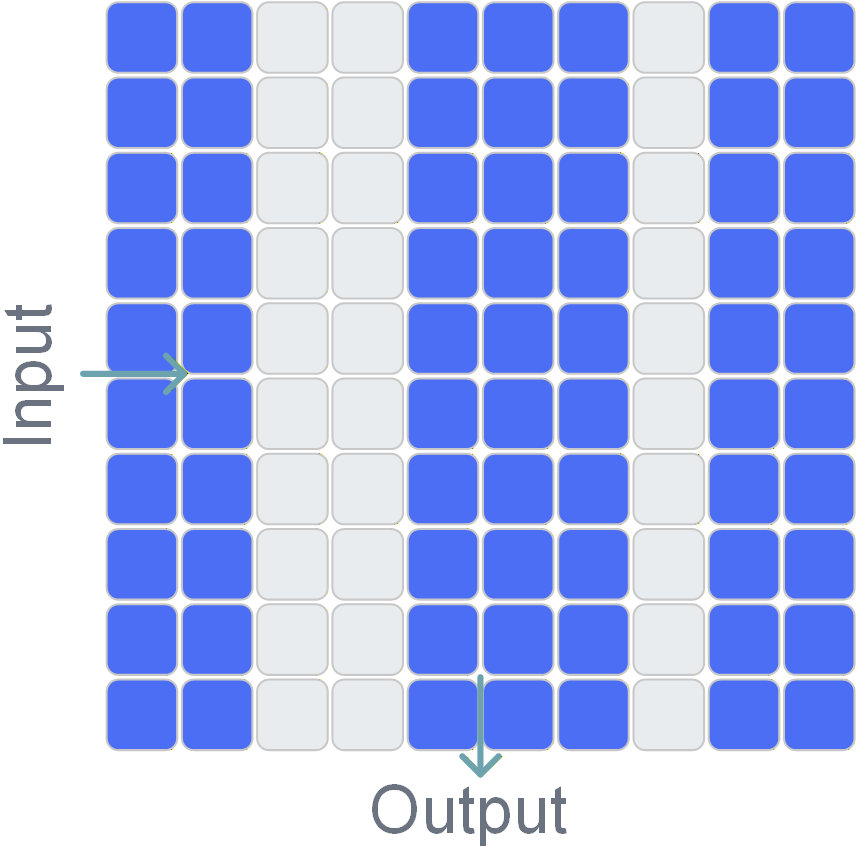}  
    \caption{State-driven sparsity.}
    \label{fig:3c}
  \end{subfigure}
  \centering
  \begin{subfigure}{0.245\linewidth}
    \centering
    \includegraphics[width=1\textwidth]{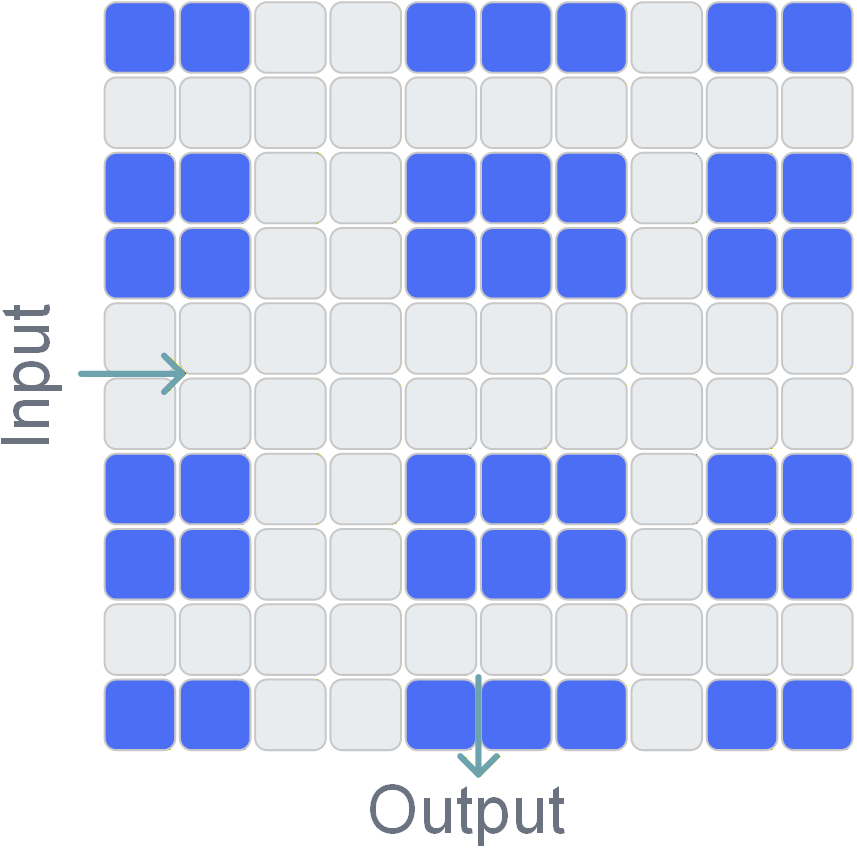}  
    \caption{Orthogonal sparsity.}
    \label{fig:3d}
  \end{subfigure}
  \caption{
    Orthogonal sparsity mechanisms in the DSN model.
    (a) In dense ANNs, all operations are active.
    (b) Traditional SNNs exploit input-driven sparsity, wherein computations involving inactive input rows are skipped.
    (c) The dormancy mechanism introduces state-driven sparsity, wherein computations involving inactive output columns are skipped.
    (d) The DSN combines both mechanisms, yielding a multiplicative reduction in the number of active synaptic operations.}
  \label{fig:orthogonal_sparsity}
\end{figure}

\subsection{Analysis of Computational Efficiency}
\label{sec:efficiency}

The primary advantage of the DSN is its ability to reduce computational cost by exploiting state-driven sparsity that is orthogonal to conventional input-driven sparsity.
Computational load is quantified using synaptic operations (SOPs).

\textbf{Input-Driven Sparsity.}
Standard SNNs gain efficiency through this presynaptic mechanism.
If the input spike rate at timestep $t$ is $\overline{R_t}$, only computations associated with the corresponding rows of the weight matrix $\mathbf{W}_n$ are performed.
The total SOPs for a network layer benefiting solely from this principle can be estimated as: $\text{SOPs}_{\text{input-driven}} = \text{MACs} \cdot \sum_{t=1}^{T} \overline{R_t}$.

\textbf{State-Driven Sparsity.}
The DSN model introduces an additional mechanism through its dormancy feature.
If the activity rate (fraction of non-dormant neurons) of the postsynaptic neurons at timestep $t$ is $\overline{D_t}$, only computations associated with the corresponding columns of the weight matrix $\mathbf{W}_n$ need to be performed.
The SOP count for a hypothetical model utilizing only this state-driven approach would be: $\text{SOPs}_{\text{state-driven}} = \text{MACs} \cdot \sum_{t=1}^{T} \overline{D_t}$.

\textbf{Orthogonal Sparsity in the DSN.}
Because the two mechanisms are orthogonal, the DSN model combines both simultaneously, allowing their effects to compound multiplicatively as shown in \cref{fig:orthogonal_sparsity}.
This yields a computational cost lower than that of either sparsity type alone: $\text{SOPs}_{\text{DSN}} = \text{MACs} \cdot \sum_{t=1}^{T} (\overline{R_t} \cdot \overline{D_t})$.
By incorporating the dormancy factor $\overline{D_t}$, the DSN can further reduce computational cost beyond the savings offered by input-driven sparsity alone.
Detailed network SOPs and energy estimates are provided in the Appendix \ref{sec:cost_calculation}.

\subsection{Inference-Time Optimization vs. Direct Training}
\label{sec:strategies}

A key feature of the DSN framework is its flexibility, offering two distinct paradigms for application: inference-time optimization (ITO) and direct training via BPTT.
The ITO paradigm acts as a plug-and-play optimization tool applied directly after training.
However, applying a fixed dormancy rule to a pre-trained network has a key limitation: early inhibition does not necessarily imply future unimportance.
For quasi-stationary inputs, ITO effectively bypasses redundant computations with limited accuracy loss.
Yet, for continuous or highly nonstationary temporal streams, dormancy may prematurely silence neurons that are crucial for processing late-arriving features, potentially causing irreversible information loss and accuracy degradation.

To address the limitations of ITO, the BPTT paradigm incorporates the dormancy mechanism directly into the end-to-end training process.
BPTT allows the network to adapt its internal representations to the dormancy constraint, reducing the likelihood that critical neurons enter dormancy prematurely and helping preserve late-arriving features.
Through the surrogate-gradient signal, the network learns a robust dormancy policy.
In this way, BPTT can help preserve task accuracy across static and dynamic temporal regimes while allowing the network to learn a task-specific balance between computational sparsity and performance.

The inference-time hardware overhead is the same for both the ITO and BPTT paradigms because both use the same dormancy mechanism.
The dormancy state is logically implemented as a 1-bit mask per neuron with a small storage footprint, allowing all subsequent 32-bit weight-memory accesses and membrane-potential updates for dormant neurons to be bypassed.
A recent neuromorphic hardware design \cite{zhou2025energy} has demonstrated that specialized synaptic processing units can leverage both spike and neuron sparsity to bypass over 98\% of invalid computations in workloads such as self-attention.
In that design, the reported synaptic energy efficiency is below 0.06 pJ/SOP, corresponding to an energy reduction of over 50\% compared with standard SNN implementations.

\begin{table}[!t]
  \caption{
    Main results on static and neuromorphic classification benchmarks.
    The DSN is evaluated under two paradigms: inference-time optimization (ITO) applied to pre-trained SNNs, and direct training with BPTT from scratch.
    Results show reductions in estimated energy ranging from 40\% to over 70\%, while maintaining accuracy comparable to or better than that of the corresponding baseline SNNs.
    Notably, the DSN outperforms state-of-the-art static pruning methods on ImageNet, delivering comparable reported cost reductions and improved accuracy.}
  \label{table:main_results}
  \small
  \renewcommand{\arraystretch}{0.55}
  \centering
  \setlength{\tabcolsep}{4pt}
  \begin{tabular}{p{4.5em} p{8.4em}<{\centering} p{6.0em}<{\centering} p{1.5em}<{\centering} p{4.5em}<{\centering} p{5em}<{\centering} p{5.5em}<{\centering}}
    \toprule[1.5pt]
    {Dataset} & {Model} & {Architecture} & \makecell{T} & \makecell{GSOPs/ \\ GFLOPs} & \makecell{Energy \\ (mJ)} & {Acc. (\%)} \\
    \midrule[1pt]
    \multirow{12.5}{*}{CIFAR10} & ANN & ResNet-18 & - & 0.56 & 2.55 & 95.93 \\
     & Transformer & 2-512 & - & 1.24 & 5.72 & 96.82 \\
     & MS-ResNet & ResNet-18 & 4 & 0.30 & 0.28 & 95.64 \\
     & SD-Transformer & 2-512 & 4 & 0.51 & 0.46 & 95.60 \\
    \cmidrule{2-7}
     & \multirow{2}{*}{\textbf{DSN (ITO)}} & ResNet-18 & 4 & \textbf{0.15} & \textbf{0.14} & \textbf{94.90} \\
      & & 2-512 & 4 & \textbf{0.31} & \textbf{0.30} & \textbf{94.29} \\
    \cmidrule{3-7}
     & \multirow{2}{*}{\textbf{DSN (BPTT)}} & ResNet-18 & 4 & \textbf{0.15} & \textbf{0.14} & \textbf{95.34} \\
      & & 2-512 & 4 & \textbf{0.30} & \textbf{0.28} & \textbf{95.59} \\
    \midrule[1pt]

    \multirow{12.5}{*}{CIFAR100} & ANN & ResNet-18 & - & 0.56 & 2.56 & 78.04 \\
     & Transformer & 2-512 & - & 1.24 & 5.72 & 81.10 \\
     & MS-ResNet & ResNet-18 & 4 & 0.32 & 0.30 & 77.32 \\
     & SD-Transformer & 2-512 & 4 & 0.63 & 0.58 & 78.40 \\
    \cmidrule{2-7}
     & \multirow{2}{*}{\textbf{DSN (ITO)}} & ResNet-18 & 4 & \textbf{0.18} & \textbf{0.16} & \textbf{76.28} \\
      & & 2-512 & 4 & \textbf{0.38} & \textbf{0.35} & \textbf{77.39} \\
    \cmidrule{3-7}
     & \multirow{2}{*}{\textbf{DSN (BPTT)}} & ResNet-18 & 4 & \textbf{0.17} & \textbf{0.16} & \textbf{76.85} \\
      & & 2-512 & 4 & \textbf{0.36} & \textbf{0.33} & \textbf{79.36} \\
    \midrule[1pt]

    \multirow{2}{*}{\makecell{DVS-\\Gesture}} & SD-Transformer & 2-256 & 16 & 1.98 & 1.82 & 96.88 \\
    \cmidrule{2-7}
    & {\textbf{DSN (BPTT)}} & 2-256 & 16 & \textbf{1.13} & \textbf{1.05} & \textbf{96.88} \\
    \midrule[1pt]

    \multirow{2}{*}{\makecell{DVS-\\CIFAR10}} & SD-Transformer & 2-256 & 16 & 1.67 & 1.51 & 78.1 \\
    \cmidrule{2-7}
    & {\textbf{DSN (BPTT)}} & 2-256 & 16 & \textbf{0.80} & \textbf{0.72} & \textbf{77.9} \\
    \midrule[1pt]

    \multirow{22}{*}{ImageNet} & ANN \cite{he2016deep} & ResNet-18 / 34 & - & 1.81 / 3.66 & 8.34 / 16.85 & 69.76 / 73.31 \\
     & Transformer \cite{dosovitskiy2020image} & 8-512 & - & 8.33 & 38.34 & 80.80 \\
     & MS-ResNet \cite{hu2024advancing} & ResNet-18 / 34 & 4 & - & 4.29 / 5.11 & 63.10 / 69.42 \\
     & SEW-ResNet \cite{fang2021deep} & ResNet-34 & 4 & 3.88 & 4.04 & 67.04 \\
     & Spikformer \cite{zhouspikformer} & 8-384 & 4 & 6.82 & 12.43 & 70.24 \\
     & SD-Transformer \cite{yao2023spike} & 8-384 & 4 & - & 3.90 & 72.28 \\
    \cmidrule{2-7}
     & AT-LIF \cite{buactivity} & ResNet-18 & 4 & $\downarrow \sim 77\%$ & - & 59.61 \\
     & DPAP \cite{han2024developmental} & ResNet-18 & - & $\downarrow \sim 51.7\%$ & - & 60.41 \\
     & Unstru. \cite{shi2024towards} & ResNet-18 & 4 & $\downarrow \sim 77\%$ & - & 61.89 \\
     & UniSparse \cite{shiunisparse} & ResNet-18 & 4 & $\downarrow \sim 78\%$ & - & 61.89 \\
     & SEENN \cite{li2023seenn} & ResNet-34 & 4 & $\downarrow \sim 55.3\%$ & - & 67.48 \\
    \cmidrule{2-7}
     & {\textbf{DSN (ITO)}} & ResNet-18 / 34 & 4 & \textbf{0.88 / 1.50} & \textbf{1.15 / 1.72} & \textbf{62.04 / 67.67} \\
    \cmidrule{3-7}
     & \multirow{2}{*}{\textbf{DSN (BPTT)}} & ResNet-18 / 34 & 4 & \textbf{0.93 / 1.67} & \textbf{1.17 / 1.83} & \textbf{64.44 / 69.63} \\
     & & 8-384 & 4 & \textbf{1.62} & \textbf{1.65} & \textbf{72.12} \\

    \bottomrule[1.5pt]
  \end{tabular}
\end{table}
\section{Experiments}
\label{sec:experiments}

\subsection{Experimental Setup}
\label{sec:setup}
The effectiveness of the proposed DSN model was evaluated across multiple benchmarks.
Static image classification performance was assessed using CIFAR10 and CIFAR100 \cite{krizhevsky2009learning}, and ImageNet \cite{russakovsky2015imagenet}.
Neuromorphic classification performance was evaluated on DVS-CIFAR10 \cite{li2017cifar10} and DVS-Gesture \cite{amir2017low}.
The experiments used two spike-driven SNN architectures, MS-ResNet and the Spike-Driven Transformer, implemented using the SpikingJelly framework \cite{fang2023spikingjelly}.
Detailed hyperparameter configurations and experimental settings are provided in the Appendix \ref{app_c2} and \cref{table:hyperparams_supp}.
The reported energy values are analytical estimates based on the hardware model described in Appendix \ref{sec:cost_calculation}, actual device-level energy and latency depend on the target hardware.

\subsection{Performance Evaluation}
\label{sec:main_results}

\textbf{Performance on Static Image Datasets.}
The results in \cref{table:main_results} show that the DSN reduces estimated energy on static image benchmarks.
A central result is that the DSN consistently reduces computational costs by substantial margins, typically exceeding 40\% and frequently reaching 50\% to 70\%, while maintaining accuracy levels comparable to the baseline SNN models.
For instance, the DSN (BPTT) applied to ResNet-18 achieves 64.44\% on ImageNet with an estimated energy cost of 1.17 mJ.
This corresponds to an estimated 73\% energy reduction relative to the 4.29 mJ baseline, accompanied by a 1.34\% improvement in accuracy.
The proposed method also outperforms recent state-of-the-art static pruning methods on ImageNet.
At comparable reported cost reductions, the listed ResNet-18 static methods show lower accuracy than the corresponding baseline, whereas the DSN achieves comparable estimated energy savings while improving upon the baseline accuracy.
These results support the effectiveness of instance-level dynamic sparsity.

Comparing the two DSN application paradigms illustrates their respective characteristics.
The ITO paradigm provides a training-free, plug-and-play optimization method.
It delivers substantial efficiency gains across all tested models but typically incurs a modest accuracy penalty.
In contrast, the direct training paradigm requires a full training process but generally preserves or improves accuracy relative to the baseline.
The results also suggest an interaction between the DSN mechanism and transformer architectures.
Applying the DSN to the SD-Transformer consistently yields substantial energy savings with limited impact on accuracy, including an accuracy increase of nearly 1\% on CIFAR100.
This suggests that the internal dynamics of transformer models may provide favorable conditions for the dormancy mechanism to remove redundant computations.

\textbf{Performance on Neuromorphic Datasets.}
The DSN was also evaluated on event-based data using DVS-Gesture and DVS-CIFAR10.
On DVS-Gesture, the DSN (BPTT) achieves the same 96.88\% accuracy as the baseline while consuming 42\% less estimated energy.
The efficiency improvement is particularly pronounced on DVS-CIFAR10, where the DSN achieves estimated energy savings of over 52\% with a 0.2\% drop in accuracy.
To examine the BPTT paradigm under longer temporal dynamics, we extended the simulation length to $T=24$ on DVS-CIFAR10.
Under this setting, the accuracy deviation from the baseline was 0.4 percentage points.
These results suggest that BPTT training can help preserve information from later timesteps over longer temporal horizons.

\textbf{Training Overhead.}
In a separate measurement, we quantified the training overhead of the DSN-BPTT by measuring training throughput and GPU memory usage for ResNet-18 on ImageNet using a single GPU with a batch size of 128.
The baseline SNN and DSN-BPTT used the same backbone architecture and training settings.
The baseline SNN achieved 742.4 images/s and used 15.3 GB of GPU memory, whereas DSN-BPTT achieved 601.6 images/s and used 20.5 GB.
These values correspond to a 19\% reduction in training throughput and a 34\% increase in GPU memory usage.
The DSN achieves reductions in estimated inference energy ranging from 40\% to over 70\%, providing a favorable trade-off for the intended deployment setting.

\subsection{Ablation Studies and Analysis}
\label{sec:ablation}

\begin{figure}[!t]
  \begin{subfigure}{0.3370\linewidth}
    \centering
    \includegraphics[width=\textwidth]{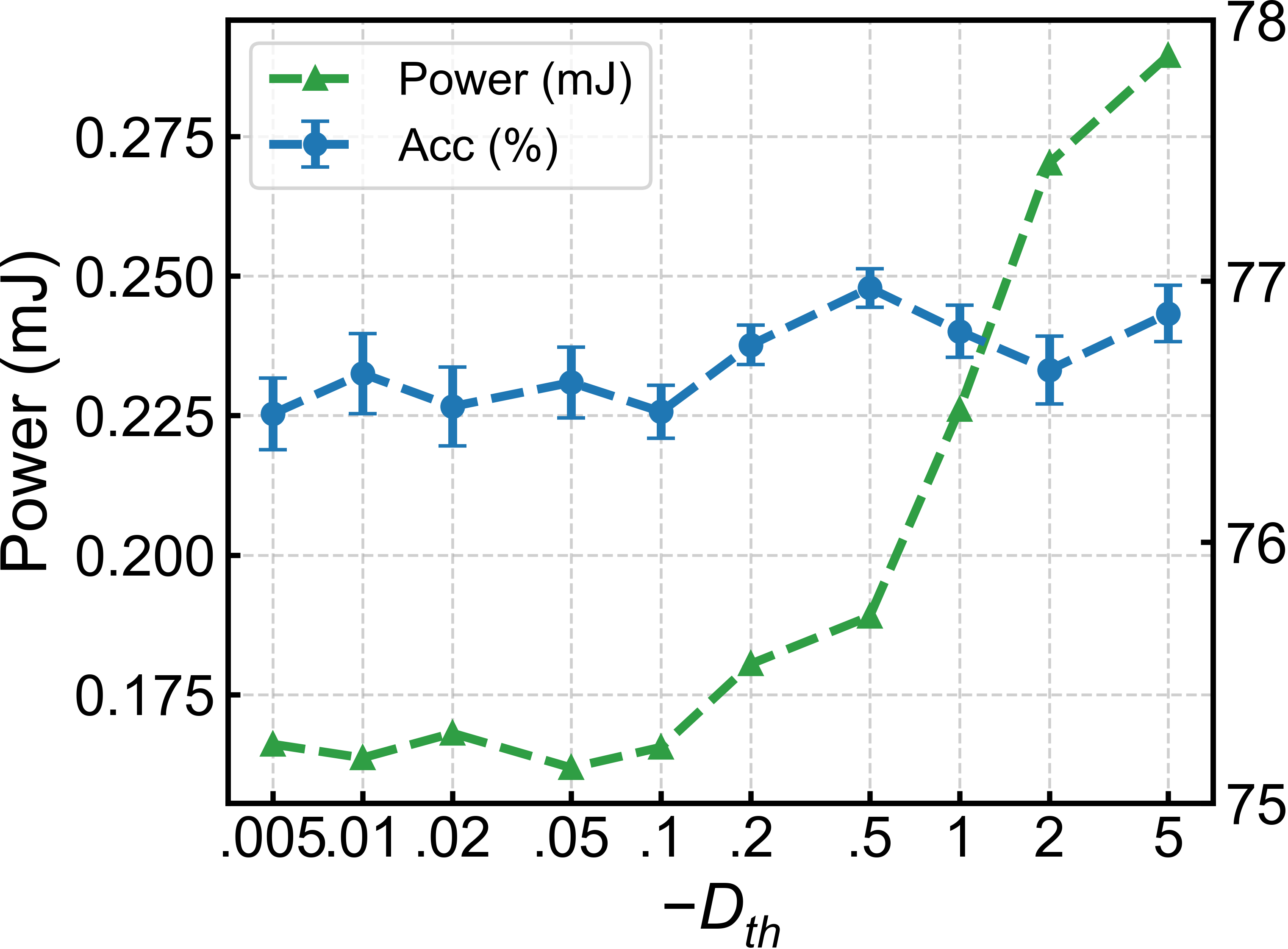}
    \caption{Dormancy threshold $D_{th}$.}
    \label{fig:ablation_dth}
  \end{subfigure}
  \centering
  \begin{subfigure}{0.2856\linewidth}
    \centering
    \includegraphics[width=\textwidth]{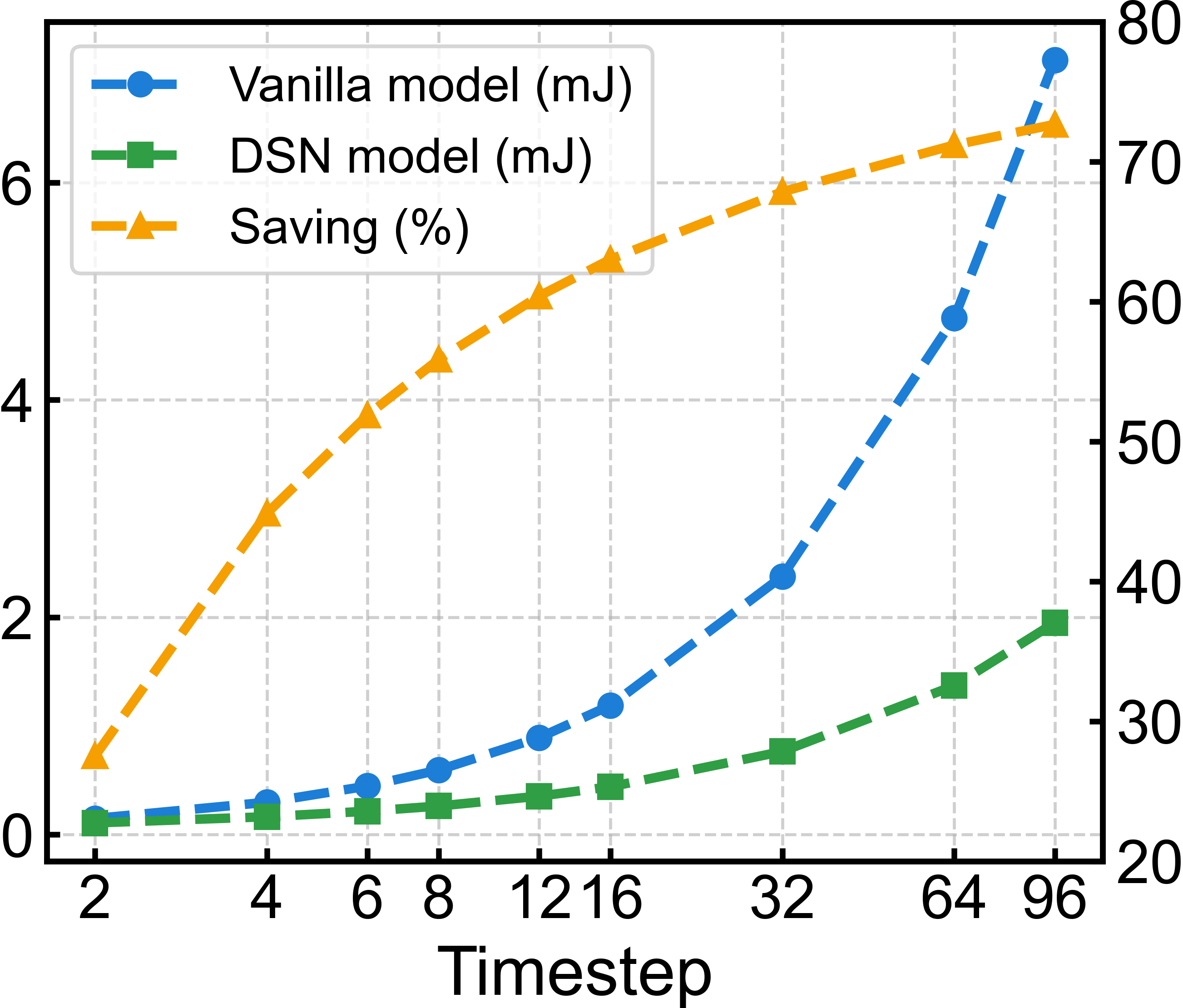}
    \caption{Temporal dynamics.}
    \label{fig:ablation_timesteps}
  \end{subfigure}
  \begin{subfigure}{0.3497\linewidth}
    \centering
    \includegraphics[width=\columnwidth]{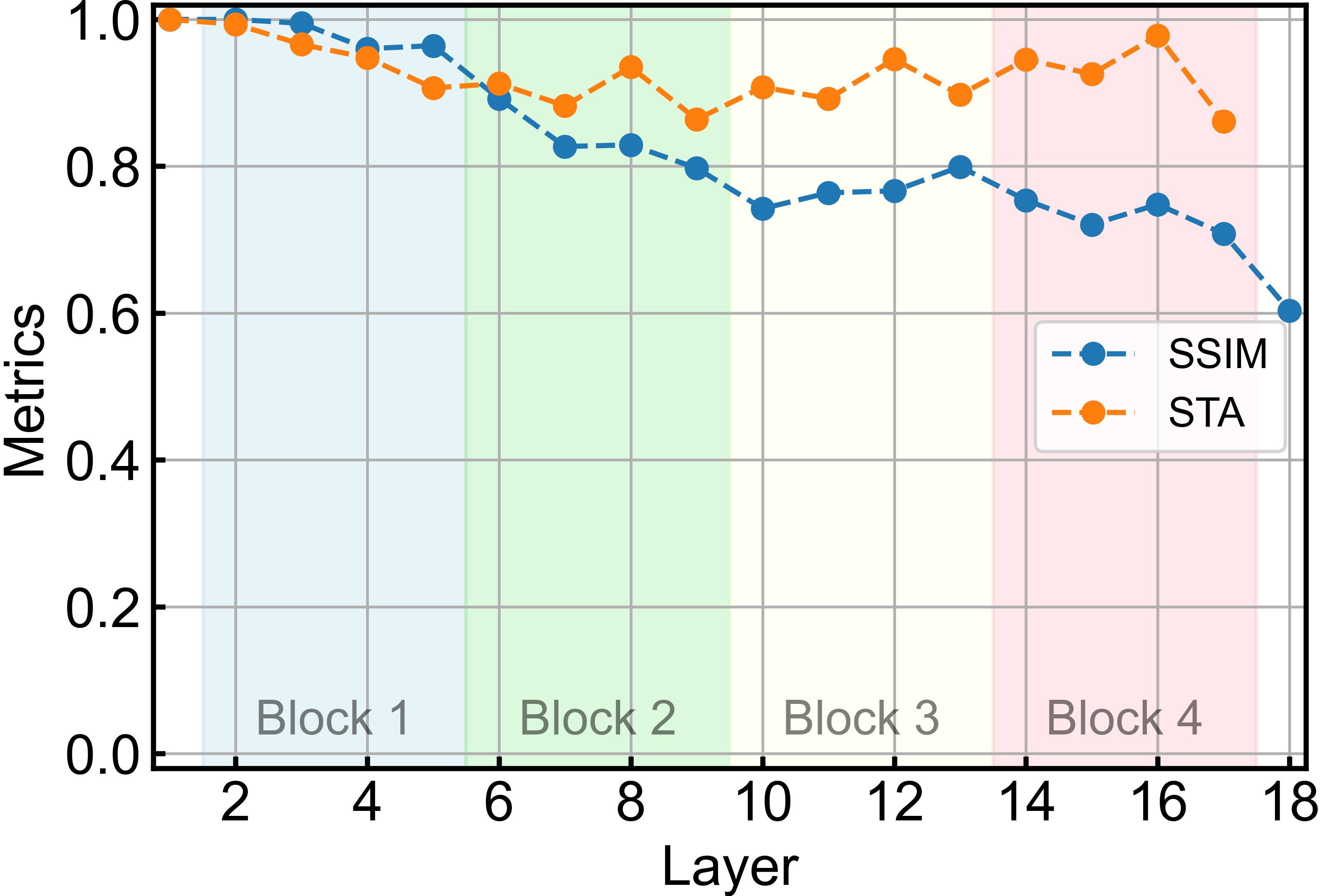}
    \caption{Impact of ITO.}
    \label{fig:ablation_ito_impact}
  \end{subfigure}
  \caption{Ablation studies on DSN dynamics.
  (a) Impact of the dormancy threshold $D_{th}$ on accuracy and energy savings evaluated across three random seeds.
  (b) Energy savings of the DSN compared with the baseline SNN as a function of the number of simulation timesteps $T$.
  (c) ITO on network representations. Layer-wise comparison of the SSIM and spike train accuracy (STA) metrics between a pre-trained baseline ResNet-18 SNN and the same network modified using the DSN ITO paradigm.}
  \label{fig:ablation}
\end{figure}

\textbf{Analysis of Dormancy Threshold.}
The dormancy threshold $D_{th}$ controls the trade-off between the aggressiveness of computational pruning and model performance.
To investigate this relationship, we evaluated ResNet-18 on CIFAR100 across three random seeds to assess run-to-run reliability.
The results, plotted in \cref{fig:ablation_dth}, reveal two primary trends.
First, accuracy remains relatively stable, with small standard deviations across the tested range of $D_{th}$.
A slight upward trend in accuracy is observable as the threshold becomes more negative, corresponding to more conservative pruning.
Second, energy consumption exhibits a distinct pattern: it remains low and relatively constant in the aggressive pruning region (where $|D_{th}| \le 0.1$) before rising sharply as the threshold becomes more conservative.
This indicates diminishing returns for energy savings when making the pruning strategy less aggressive than $D_{th} \approx -0.1$.
Based on this analysis, $D_{th}=-0.05$ was selected for static datasets, providing substantial energy savings near the low-energy plateau while preserving accuracy in the tested range.
For neuromorphic datasets, a more conservative threshold of $D_{th}=-0.5$ was adopted to prioritize performance stability under dynamic event-based data streams.

\textbf{Multi-seed Evaluation.}
To assess run-to-run reliability, we repeated the BPTT-trained DSN configurations corresponding to the main results using three random seeds.
CIFAR10, CIFAR100, and ImageNet use ResNet-18, whereas DVS-Gesture and DVS-CIFAR10 use the Spike-Driven Transformer.
The mean and standard deviation of the classification accuracy are reported in \cref{table:multi_seed}.
The standard deviations remain small across all evaluated datasets, indicating limited run-to-run variation.

\textbf{Impact of Simulation Timesteps on Energy Savings.}
The efficiency benefit provided by the DSN increases with the number of timesteps, as more neurons are progressively identified as deeply inhibited and transition to the dormant state.
This benefit was evaluated by comparing the energy consumption of a DSN-equipped ResNet-18 with that of its baseline SNN on CIFAR100 over varying simulation lengths $T$.
The results are plotted in \cref{fig:ablation_timesteps}.
Even with a short simulation window of $T=2$, the DSN achieves energy savings of approximately 30\%.
At $T=4$, it achieves energy savings of 45\%.
These results indicate that the DSN can provide computational reductions in low-latency regimes and may be compatible with early-exit or dynamic-timestep strategies.
Furthermore, this efficiency gap widens considerably as the simulation duration increases, growing to over 63\% at $T=16$ and exceeding 73\% by $T=96$.
This demonstrates that the dormancy mechanism becomes increasingly effective over longer sequences.

\textbf{Impact of ITO on Network Representations.}
To further understand the mechanisms underlying the effectiveness of the ITO paradigm, its impact on the internal representations learned by the network was analyzed.
Two metrics were employed for this layer-wise comparison using a ResNet-18 SNN pretrained on ImageNet.
First, the structural similarity index measure (SSIM) \cite{wang2004image} was used to assess the similarity between the pre-activation feature maps of the original SNN and the DSN-modified network.
High SSIM values indicate that the core spatial features extracted by the layers are well preserved after applying ITO.
Second, spike train accuracy (STA) was used to measure the fidelity of the output spike trains generated by each layer after the ITO modification compared to the original SNN.
This metric is calculated as the average fraction of exact spike matches over all neurons and timesteps between the original ($S$) and DSN-modified ($\hat{S}$) spike trains for a given layer $L$:
\begin{equation}
	\text{STA}_L = \frac{1}{N_L \cdot T} \sum_{i=1}^{N_L} \sum_{t=1}^{T} \mathbb{I}(S_{L,i}^t = \hat{S}_{L,i}^t)
	\label{eq:sta}
\end{equation}
where $N_L$ is the number of neurons in layer $L$, and $\mathbb{I}(\cdot)$ is the indicator function.
\Cref{fig:ablation_ito_impact} presents the layer-wise SSIM and STA curves.
The consistently high STA values indicate that the dormancy mechanism has little effect on the spike-firing patterns of the original network, supporting the claimed orthogonality between state-driven and input-driven sparsity mechanisms.

\begin{table}[!t]
  \caption{
    Three-seed accuracy results for the BPTT-trained DSN.
    Values are reported as mean $\pm$ standard deviation over three runs.
  }
  \label{table:multi_seed}
  \small
  \centering
  \begin{tabular}{lccccc}
    \toprule
    & CIFAR10 & CIFAR100 & ImageNet & DVS-Gesture & DVS-CIFAR10 \\
    \midrule
    Accuracy (\%) &
    $95.30 \pm 0.10$ &
    $76.61 \pm 0.13$ &
    $64.36 \pm 0.19$ &
    $96.65 \pm 0.40$ &
    $77.83 \pm 0.12$ \\
    \bottomrule
  \end{tabular}
\end{table}

\textbf{Analysis of Neuronal Computational Utilization.}
To provide deeper mechanistic insight into the efficiency gains afforded by the DSN, another metric was introduced to quantify the computational state of neurons.
Let $\overline{R}$ denote the average firing rate, and let $\overline{D}$ denote the average fraction of non-dormant neurons.
Building upon these quantities, the neuronal computational utilization (NCU) metric measures the fraction of neurons that either fire or have entered the dormant state:
\begin{equation}
  \text{NCU} = \overline{R} + (1 - \overline{D})
  \label{eq:ncu}
\end{equation}
An ideal NCU of 1 signifies that no computational resources are spent on neurons lingering in the subthreshold active state (neither firing nor dormant).
Recalling the initial motivation, which identified approximately 44\% of neurons residing in a deeply inhibited, computationally redundant state, the NCU metric quantifies how much of this computational redundancy is addressed by the DSN.
Layer-wise analysis of a BPTT-trained model on ImageNet supports this interpretation.
The empirical results suggest that the DSN does not achieve computational efficiency by indiscriminately suppressing the active firing rate $\overline{R}$.
Instead, the energy savings are primarily associated with an increased dormancy fraction $(1-\overline{D})$ among computationally redundant neurons.
Across most network layers, the NCU surpasses 70\%.
This indicates that the DSN converts a large portion of the identified deeply inhibited population into the dormant state, where subsequent computations are bypassed, thereby addressing the computational redundancy identified in the introduction.
The remaining NCU gap largely corresponds to neurons in the subthreshold active region, whose optimization falls outside the scope of this specific dormancy mechanism but suggests avenues for future research.

\textbf{Jaccard Similarity of Dormancy Masks.}
To distinguish the DSN from conventional static pruning techniques, the Jaccard similarity of dormant masks across diverse input samples was analyzed.
Evaluating 10,000 ImageNet image pairs on a ResNet-34 model revealed high heterogeneity in the dormancy patterns, with an average mask overlap of 0.40.
Furthermore, more than 12 layers showed minimal overlap below 0.20.
In contrast, for static pruning methods, the Jaccard similarity is fixed at 1 across input samples.
This variation supports the conclusion that the DSN leverages instance-level rather than dataset-level redundancy and is consistent with its advantage over static pruning methods.

\textbf{Validation of the Surrogate Gradient for Dormancy.}
While applying surrogate gradients is standard practice for handling the non-differentiable spike-firing mechanism in conventional SNN training, the DSN introduces an additional non-differentiable component via the dormancy gate.
To validate the effectiveness of the proposed BPTT methodology specifically tailored for the DSN, an ablation study was conducted to compare the use of a smooth SG for the dormancy gate update against a simpler straight-through estimator (STE) \cite{courbariaux2015binaryconnect}.
The results show higher accuracy for the SG-based dormancy gate on the evaluated benchmarks.
For instance, on the CIFAR10 and CIFAR100 datasets, the SG approach improves accuracy by 0.30\% and 0.63\%, respectively.
More notably, on the large-scale ImageNet dataset, the SG approach achieves 64.44\% accuracy, exceeding the STE baseline by 1.57\%.
This suggests that a smooth gradient signal for the dormancy gate helps the network learn a robust dormancy policy during end-to-end training.

\section{Limitations and Discussion}
\label{sec:limitation}

The proposed DSN model demonstrates potential for improving SNN efficiency by directly targeting the computational overhead associated with persistently inhibited neurons.
However, the applicability of the DSN framework depends on the neural coding scheme.
Specifically, the dormancy mechanism relies on information redundancy and is particularly compatible with rate coding and multi-spike architectures.
Conversely, it may be less suitable for low-redundancy coding paradigms, such as Time-to-First-Spike (TTFS) coding.
In TTFS schemes, information is encoded in the precise timing of a single spike, leaving limited informational redundancy for the dormancy gate to exploit without risking performance degradation.
This limitation does not preclude the practical utility of the proposed method.
Networks using TTFS coding generally operate with high sparsity and low baseline computational overhead.
Consequently, both the available margin and the practical need for further computational reduction in such highly sparse regimes may be limited.

Beyond efficiency gains, this work suggests a direction for SNN optimization.
By shifting the focus from presynaptic input sparsity to postsynaptic state-driven sparsity, the DSN provides a complementary framework for exploiting redundancy in continuous neuronal state updates.
Furthermore, insights from the neuronal computational utilization analysis suggest an avenue for future research.
While the DSN bypasses computations for deeply inhibited neurons, developing fine-grained optimization strategies for those remaining in the subthreshold active region could improve computational utilization and help explore improved energy-accuracy trade-offs in neuromorphic systems.

\section{Conclusion}
\label{sec:conclusion}

This work addressed the substantial but often overlooked computational overhead in SNNs arising from continuous subthreshold updates of deeply inhibited neurons.
A quantitative analysis characterized the magnitude of this inefficiency, motivating the development of the Dormant Spiking Neuron (DSN).
The DSN integrates a mechanism that allows neurons to enter a permanent dormant state based on their membrane potentials, thereby bypassing redundant computations at the instance level.
To improve robustness on temporal streams, this state-driven mechanism is trained end-to-end via backpropagation through time using surrogate gradients.
Extensive evaluations demonstrated that the DSN framework reduces estimated energy consumption by 40\% to over 70\% across static and neuromorphic datasets and network architectures.
These efficiency gains were achieved while maintaining competitive task accuracy relative to the baselines.
On ImageNet, the DSN outperformed the evaluated static pruning methods at comparable reported cost reductions.
By providing a trainable, state-driven sparsity mechanism orthogonal to conventional presynaptic input-driven sparsity, the DSN offers an effective approach for improving neuromorphic computing efficiency.

\section*{Acknowledgements}
This work was supported in part by the Brain Science and Brain-like Intelligence Technology-National Science and Technology Major Project 2022ZD0209700 and in part by Sichuan Provincial Industrial Development Fund under Grant 2025JB04 and in part by Shenzhen Natural Science Foundation under Grant JCYJ20250604180428038.

{
\small
\bibliographystyle{unsrt}
\bibliography{main}
}

\newpage
\appendix

\section{Detailed BPTT Derivation for the DSN}
\label{app:bptt_derivation}

\subsection{Forward Dynamics and Gradient Notations}

In this section, we provide the rigorous, step-by-step derivations for the BPTT algorithm tailored specifically to the Dormant Spiking Neuron (DSN).
We first reiterate the forward propagation dynamics of the proposed DSN model and establish the notations utilized throughout the backward pass.
The dynamics of a DSN in layer $n$ at timestep $t$ are governed by the following four stages:

\textbf{1. Dormancy Gate Update.}
The binary activity state $D_n^t$ is dynamically updated based on the membrane potential from the preceding timestep $V_n^{t-1}$ and the predefined dormancy threshold $D_{th}$:
\begin{equation} \label{eq:app_D}
D_n^t = D_n^{t-1} \cdot \Theta(V_n^{t-1} - D_{th})
\end{equation}
\textbf{2. Potential Integration.}
The neuron integrates incoming spikes $S_{n-1}^t$ from the preceding layer, weighted by $\mathbf{W}_n$, with the decayed membrane potential from the previous timestep:
\begin{equation} \label{eq:app_H}
H_n^t = f(V_n^{t-1}) + \mathbf{W}_n S_{n-1}^t
\end{equation}
\textbf{3. Gated Spike Firing.}
The output spike $S_n^t$ is generated only if the integrated potential $H_n^t$ exceeds the firing threshold $V_{th}$ and the neuron remains in the active state ($D_n^t=1$):
\begin{equation} \label{eq:app_S}
S_n^t = D_n^t \cdot \Theta(H_n^t - V_{th})
\end{equation}
\textbf{4. Potential Update \& Reset.}
The membrane potential is subsequently updated for the next timestep, incorporating a hard reset mechanism for neurons that have fired:
\begin{equation} \label{eq:app_V}
V_n^t = H_n^t (1 - S_n^t) + V_{\text{reset}} S_n^t
\end{equation}

Here, $f(V)$ represents the membrane potential decay function (e.g., $f(V) = \tau^{-1} \cdot V$ for a standard LIF neuron model).
To formalize the BPTT derivation, we define the gradients of the total loss $L$ with respect to the key network variables at layer $n$ and timestep $t$ as follows:
\begin{itemize}
\item $\delta_{H,n}^t \equiv \frac{\partial L}{\partial H_n^t}$: Gradient with respect to the integrated potential $H$.
\item $\delta_{S,n}^t \equiv \frac{\partial L}{\partial S_n^t}$: Gradient with respect to the spike output $S$.
\item $\delta_{V,n}^t \equiv \frac{\partial L}{\partial V_n^t}$: Gradient with respect to the updated membrane potential $V$.
\item $\delta_{D,n}^t \equiv \frac{\partial L}{\partial D_n^t}$: Gradient with respect to the dormancy gate state $D$.
\end{itemize}

Because the Heaviside step function $\Theta(\cdot)$ in \cref{eq:app_D,eq:app_S} is non-differentiable, its exact derivative is substituted with a continuous surrogate gradient function $\sigma'(\cdot)$ during the backward pass:
\begin{itemize}
\item $\frac{\partial \Theta(x)}{\partial x} := \sigma_d'(x)$ for the dormancy gate update in \cref{eq:app_D}.
\item $\frac{\partial \Theta(x)}{\partial x} := \sigma_s'(x)$ for spike generation in \cref{eq:app_S}.
\end{itemize}

\subsection{Backward Pass Gradient Derivation}
The ultimate objective of the BPTT process is to compute the gradient of the loss with respect to the network weights, $\frac{\partial L}{\partial \mathbf{W}_n}$.
Using the multivariate chain rule, this gradient is accumulated over the total simulation timesteps $T$, relying crucially on the intermediate gradient $\delta_{H,n}^t$:
\begin{equation} \label{eq:app_W}
\frac{\partial L}{\partial \mathbf{W}_n} = \sum_{t=1}^{T} \frac{\partial L}{\partial H_n^t} \frac{\partial H_n^t}{\partial \mathbf{W}_n} = \sum_{t=1}^{T} \delta_{H,n}^t (S_{n-1}^t)^\top
\end{equation}
The core task is to unroll the recurrence relations for the defined error terms ($\delta_{H,n}^t$, $\delta_{S,n}^t$, $\delta_{V,n}^t$, $\delta_{D,n}^t$) by propagating the error signals backward through both network depth and time.

\subsubsection{Step 1: Deriving $\delta_{H,n}^t$}
At timestep $t$, $H_n^t$ independently influences the current spike generation $S_n^t$ (\cref{eq:app_S}) and the subsequent reset potential $V_n^t$ (\cref{eq:app_V}).
The gradient $\delta_{H,n}^t$ accumulates from these two paths:
\[
\delta_{H,n}^t = \frac{\partial L}{\partial H_n^t} = \underbrace{\frac{\partial L}{\partial S_n^t} \frac{\partial S_n^t}{\partial H_n^t}}_{\text{Firing Path}} + \underbrace{\frac{\partial L}{\partial V_n^t} \frac{\partial V_n^t}{\partial H_n^t}}_{\text{Reset Path}}
\]
To ensure training stability and mitigate potential gradient anomalies, it is standard practice in SNNs to detach the spike tensor $S_n^t$ within the reset equation (\cref{eq:app_V}) from the computational graph during the backward pass (treating it as a constant).
This widely adopted approximation simplifies the reset path gradient to $\frac{\partial V_n^t}{\partial H_n^t} = 1 - S_n^t$. Consequently, we obtain:
\begin{equation} \label{eq:app_G1}
\boxed{
\delta_{H,n}^t = \delta_{S,n}^t \cdot \left[ D_n^t \cdot \sigma_s'(H_n^t - V_{th}) \right] + \delta_{V,n}^t \cdot (1 - S_n^t)
}
\end{equation}

\subsubsection{Step 2: Deriving $\delta_{S,n}^t$}
The spike output $S_n^t$ propagates forward to influence the integrated potential of the next layer $H_{n+1}^t$ (\cref{eq:app_H}) and locally affects the intra-neuron reset potential $V_n^t$ (\cref{eq:app_V}).
\[
\delta_{S,n}^t = \frac{\partial L}{\partial S_n^t} = \underbrace{\frac{\partial L}{\partial H_{n+1}^t} \frac{\partial H_{n+1}^t}{\partial S_n^t}}_{\text{Spatial Path}} + \underbrace{\frac{\partial L}{\partial V_n^t} \frac{\partial V_n^t}{\partial S_n^t}}_{\text{Reset Path}}
\]
As established in Step 1, the reset path dependency is detached, rendering $\frac{\partial V_n^t}{\partial S_n^t} = 0$.
Thus, the error gradient flows exclusively backward through the spatial path from the subsequent layer:
\begin{equation} \label{eq:app_G2}
\boxed{
\delta_{S,n}^t = \mathbf{W}_{n+1}^\top \delta_{H,n+1}^t
}
\end{equation}

\subsubsection{Step 3: Deriving $\delta_{V,n}^t$}
The membrane potential $V_n^t$ explicitly carries the temporal state to the next timestep $t+1$, jointly influencing the future dormancy gate $D_n^{t+1}$ (\cref{eq:app_D}) and the future integrated potential $H_n^{t+1}$ (\cref{eq:app_H}).
\[
\delta_{V,n}^t = \frac{\partial L}{\partial V_n^t} = \underbrace{\frac{\partial L}{\partial D_n^{t+1}} \frac{\partial D_n^{t+1}}{\partial V_n^t} + \frac{\partial L}{\partial H_n^{t+1}} \frac{\partial H_n^{t+1}}{\partial V_n^t}}_{\text{Temporal Path}}
\]
Evaluating these partial derivatives yields the temporal gradient relation:
\begin{equation} \label{eq:app_G3}
\boxed{
\delta_{V,n}^t = \delta_{D,n}^{t+1} \cdot \left[ D_n^t \cdot \sigma_d'(V_n^t - D_{th}) \right] + \delta_{H,n}^{t+1} \cdot f'(V_n^t)
}
\end{equation}
Specifically, for the LIF neuron defined by $f(V_n^{t}) = \tau^{-1} \cdot V_n^{t}$, the derivative is $f'(V_n^t) = \tau^{-1}$. Conversely, for the IF neuron where $f(V_n^{t}) = V_n^{t}$, the derivative evaluates to $f'(V_n^t) = 1$.

\subsubsection{Step 4: Deriving $\delta_{D,n}^t$}
The dormancy gate state $D_n^t$ structurally governs both its own temporal continuation $D_n^{t+1}$ (\cref{eq:app_D}) and the immediate spatial spike generation $S_n^t$ (\cref{eq:app_S}).
\[
\delta_{D,n}^t = \frac{\partial L}{\partial D_n^t} = \underbrace{\frac{\partial L}{\partial D_n^{t+1}} \frac{\partial D_n^{t+1}}{\partial D_n^t}}_{\text{Temporal Path}} + \underbrace{\frac{\partial L}{\partial S_n^t} \frac{\partial S_n^t}{\partial D_n^t}}_{\text{Firing Path}}
\]
Substituting the derivatives, we formalize the gradient for the dormancy mechanism:
\begin{equation} \label{eq:app_G4}
\boxed{
\delta_{D,n}^t = \delta_{D,n}^{t+1} \cdot \Theta(V_n^t - D_{th}) + \delta_{S,n}^t \cdot \Theta(H_n^t - V_{th})
}
\end{equation}

\subsection{The Complete BPTT Algorithm}
By consolidating the derived recurrence relations, we establish the complete, implementable algorithmic workflow for the DSN's backward pass.
\begin{enumerate}
\item \textbf{Initialization:}
\begin{itemize}
  \item For all layers $n \in \{1, \dots, N\}$, initialize the temporal error terms originating from the hypothetical future step $T+1$ to zero: $\delta_{H,n}^{T+1} = 0$ and $\delta_{D,n}^{T+1} = 0$.
  \item For all layers $n$, initialize the weight gradient accumulators: $\Delta \mathbf{W}_n = \mathbf{0}$.
  \item For all timesteps $t \in \{1, \dots, T\}$, compute the initial error gradient at the final output layer $N$, $\delta_{S, N}^t = \frac{\partial L}{\partial S_N^t}$, derived directly from the task loss function $L$.
\end{itemize}
\item \textbf{Iterative Gradient Unrolling:}
For each timestep $t$ from $T$ down to $1$:
\begin{itemize}
    \item For each layer $n$, iterating backward from $N$ down to $1$:
  \begin{enumerate}
      \item \textbf{Calculate Spatial Spike Error $\delta_{S,n}^t$:} If $n < N$, compute via \cref{eq:app_G2}.
      \item \textbf{Calculate Dormancy Gate Error $\delta_{D,n}^t$:} Compute via \cref{eq:app_G4}, utilizing $\delta_{D,n}^{t+1}$ (from the previous temporal loop) and $\delta_{S,n}^t$.
      \item \textbf{Calculate Membrane Potential Error $\delta_{V,n}^t$:} Compute via \cref{eq:app_G3}, utilizing $\delta_{D,n}^{t+1}$ and $\delta_{H,n}^{t+1}$.
      \item \textbf{Calculate Integrated Potential Error $\delta_{H,n}^t$:} Compute via \cref{eq:app_G1}, bridging $\delta_{S,n}^t$ and $\delta_{V,n}^t$.
      \item \textbf{Accumulate Weight Gradient:} 
      $\Delta \mathbf{W}_n \leftarrow \Delta \mathbf{W}_n + \delta_{H,n}^t (S_{n-1}^t)^\top$
  \end{enumerate}
  \end{itemize}
  \item \textbf{Final Weight Update:}
Upon completion of the loops, the exact aggregated gradient for the optimizer is defined as $\frac{\partial L}{\partial \mathbf{W}_n} = \Delta \mathbf{W}_n$.
\end{enumerate}
This rigorous formulation ensures that the DSN, complete with its dynamic state-driven sparsity gating, can be trained end-to-end using modern automatic differentiation frameworks.

\section{Computational and Energy Cost Calculation}
\label{sec:cost_calculation}

\subsection{FLOPs Calculation for Standard Architectures}
The computational cost of conventional ANNs is commonly reported in FLOPs.
For consistency with our energy accounting, one MAC is counted as one FLOP.

For a standard convolutional layer parameterized by $C_{in}$ input channels, $C_{out}$ output channels, a spatial kernel size of $K \times K$, and producing an output feature map of dimensions $H_{out} \times W_{out}$, the total number of dense MACs is formulated as:
$\text{MACs}_{\text{Conv}} = H_{out} \cdot W_{out} \cdot K \cdot K \cdot C_{in} \cdot C_{out}$.

Similarly, for a standard fully connected (linear) layer bridging $N_{in}$ input dimensions to $N_{out}$ output dimensions, the total MAC count is:
$\text{MACs}_{\text{MLP}} = N_{in} \cdot N_{out}$.
The aggregate FLOP count for an entire ANN is simply the algebraic sum of the individual MACs across all constituent layers.

\subsection{SOPs Calculation for Spiking Architectures}
In stark contrast to ANNs, the computational burden of spiking layers is quantified via SOPs.
Because spike activations are binary ($S \in \{0,1\}$), the heavy floating-point multiplication in an SNN is bypassed, leaving only an event-driven conditional accumulation (AC), which is hardware-efficient.
Consequently, the total SOP count for a spiking layer $n$ is merely a fraction of its dense ANN counterpart ($\text{MACs}_{n}$), fundamentally dictated by the layer's temporal sparsity.

\paragraph{Standard SNNs.}
Traditional SNNs derive their computational efficiency predominantly from presynaptic input-driven sparsity.
A synaptic operation is executed strictly when an active input spike is emitted from the preceding layer $n-1$.
Thus, the total SOPs for a baseline spiking layer $n$ equate to its dense MACs scaled by the average temporal firing rate of the input spikes, denoted as $\overline{R_{n-1}^t}$:
\[
\text{SOPs}_{\text{SNN}, n} = \text{MACs}_{n} \cdot \sum_{t=1}^{T} \overline{R_{n-1}^t}
\]

\paragraph{The DSN Model.}
The proposed DSN extends this baseline by introducing a second, orthogonal mechanism: state-driven sparsity.
Within the DSN framework, an operation at layer $n$ is executed only if two conditions are simultaneously met: an active input spike arrives from layer $n-1$, and the designated receiving postsynaptic neuron in layer $n$ remains in a non-dormant state.
By exploiting this joint filtering effect, the SOP count for a DSN layer $n$ is reduced multiplicatively:
\[
\text{SOPs}_{\text{DSN}, n} = \text{MACs}_{n} \cdot \sum_{t=1}^{T} (\overline{R_{n-1}^t} \cdot \overline{D_{n}^t})
\]
Here, the two sparsity factors are defined as follows:\\
$\overline{R_{n-1}^t}$:
The presynaptic activity rate, represented by the mean of the binary spike matrix $S_{n-1}^t$.\\
$\overline{D_{n}^t}$:
The fraction of active (non-dormant) neurons in the postsynaptic layer $n$, represented by the mean of the binary dormancy gate matrix $D_n^t$.

\begin{table}[t]
    \renewcommand{\arraystretch}{1.6}
    \small
    \setlength{\tabcolsep}{0.5pt}
    \caption{Energy calculation example for a Spike-Driven Transformer architecture.}
    \label{table:transformer_sops}
    \centering
    \begin{tabular}{c|c|c|c}
        \toprule[1.5pt]
            & & \makecell{Vanilla \\ Spike-Driven Transformer} & \makecell{Spike-Driven Transformer \\ with DSN} \\
        \midrule[1pt]
        \multirow{2.2}{*}{SPS} & First Conv & $E_{mac} \cdot \text{MACs}_{Conv_1}$ & $E_{mac} \cdot \text{MACs}_{Conv_1}$ \\
            & Other Conv & $E_{ac} \cdot \text{MACs}_{Conv_n} \cdot \sum_{t=1}^{T} \overline{R_{n-1}^t}$ & $E_{ac} \cdot \text{MACs}_{Conv_n} \cdot \sum_{t=1}^{T} (\overline{R_{n-1}^t} \cdot \overline{D_{n}^t})$ \\
        \midrule[1pt]
        \multirow{5.8}{*}{Self-Attention} & $Q$ & $E_{ac} \cdot ND^2 \cdot \sum_{t=1}^{T} \overline{R_{in}^t}$ & $E_{ac} \cdot ND^2 \cdot \sum_{t=1}^{T} (\overline{R_{in}^t} \cdot \overline{D_{Q}^t})$ \\
         & $K$ & $E_{ac} \cdot ND^2 \cdot \sum_{t=1}^{T} \overline{R_{in}^t}$ & $E_{ac} \cdot ND^2 \cdot \sum_{t=1}^{T} (\overline{R_{in}^t} \cdot \overline{D_{K}^t})$ \\
         & $V$ & $E_{ac} \cdot ND^2 \cdot \sum_{t=1}^{T} \overline{R_{in}^t}$ & $E_{ac} \cdot ND^2 \cdot \sum_{t=1}^{T} (\overline{R_{in}^t} \cdot \overline{D_{V}^t})$ \\
         & $f(Q,K,V)$ & $E_{ac} \cdot ND \cdot \sum_{t=1}^{T} \overline{R_{KV}^t}$ & $E_{ac} \cdot ND \cdot \sum_{t=1}^{T} \overline{R_{KV}^t}$ \\
         & Linear 0 & $E_{ac} \cdot ND^2 \cdot \sum_{t=1}^{T} \overline{R_{QKV}^t}$ & $E_{ac} \cdot ND^2 \cdot \sum_{t=1}^{T} (\overline{R_{QKV}^t} \cdot \overline{D_{MLP_0}^t})$ \\
        \midrule[1pt]
        \multirow{2.2}{*}{MLP} & Linear 1 & $E_{ac} \cdot 4ND^2 \cdot \sum_{t=1}^{T} \overline{R_{MLP_0}^t}$ & $E_{ac} \cdot 4ND^2 \cdot \sum_{t=1}^{T} (\overline{R_{MLP_0}^t} \cdot \overline{D_{MLP_1}^t})$ \\

         & Linear 2 & $E_{ac} \cdot 4ND^2 \cdot \sum_{t=1}^{T} \overline{R_{MLP_1}^t}$ & $E_{ac} \cdot 4ND^2 \cdot \sum_{t=1}^{T} (\overline{R_{MLP_1}^t} \cdot \overline{D_{MLP_2}^t})$ \\
        \bottomrule[1.5pt]
    \end{tabular}%
\end{table}

\subsection{Energy Consumption Model}
To map the aforementioned computational counts to energy estimates, we adopt the standard energy estimation framework established for a 45nm CMOS process node.
The theoretical energy costs for 32-bit operations are defined as follows:
$E_{mac} = 4.6 \text{ pJ}$ for one dense multiply-and-accumulate (MAC) operation;
$E_{ac} = 0.9 \text{ pJ}$ for one sparse accumulate (AC) operation.

The total inference energy for an ANN is computed by integrating the MAC energy costs across all $N$ layers of the network:
$\text{Energy}_{\text{ANN}} = E_{mac} \cdot \sum_{n=1}^{N} \text{MACs}_n$.

The energy calculation for spiking architectures is distinct due to the dominance of low-power AC operations.
Following standard SNN encoding protocols, the first layer ($n=1$) operates as an analog-to-spike encoder, executing dense MACs, whereas all subsequent layers ($n > 1$) operate purely in the spike domain, executing sparse ACs.
Consequently, the total energy consumption is the hybrid sum of the first layer's MAC energy and the deep layers' SOP (AC) energy:
\begin{equation}
\text{Energy}_{\text{SNN}} = \underbrace{E_{mac} \cdot \text{MACs}_1}_{\text{First Layer (ANN)}} + \underbrace{E_{ac} \cdot \sum_{n=2}^{N} \text{SOPs}_n}_{\text{Spiking Layers}}
\end{equation}
In this formulation, the term $\text{SOPs}_n$ is substituted with $\text{SOPs}_{\text{SNN}, n}$ when evaluating a vanilla SNN baseline, and replaced by the highly reduced $\text{SOPs}_{\text{DSN}, n}$ when evaluating the proposed DSN model.

\subsubsection{Spike-Driven Transformer Energy Analysis}
To provide a concrete architectural example, \cref{table:transformer_sops} details the energy calculations for the components of a Spike-Driven Transformer block.
This illustrates the precise locations where the DSN injects state-driven sparsity compared to the vanilla SNN baseline.
The derivations are formulated based on the sequence length $N$, the embedding channel dimension $D$, and the MLP expansion ratio $4D$.

For clarity, the respective spike sparsity rates are defined as follows:
$\overline{R_{in}}$ denotes the input spike rate entering the transformer block;
$\overline{R_{KV}}$ represents the activation sparsity following the Key-Value (KV) matrix summation;
$\overline{R_{QKV}}$ signifies the spike sparsity emerging from the core attention computation $f(Q,K,V)$;
and $\overline{R_{MLP_n}}$ denotes the spike rate propagating from the respective linear layers within the MLP block.
Correspondingly, the DSN's non-dormant activity factors $\overline{D}$ are applied to the postsynaptic neurons in each internal module, enabling multiplicative reductions in computational cost throughout the self-attention mechanism.

\begin{figure}[!t]
  \begin{subfigure}{0.32\linewidth}
    \centering
    \includegraphics[width=\textwidth]{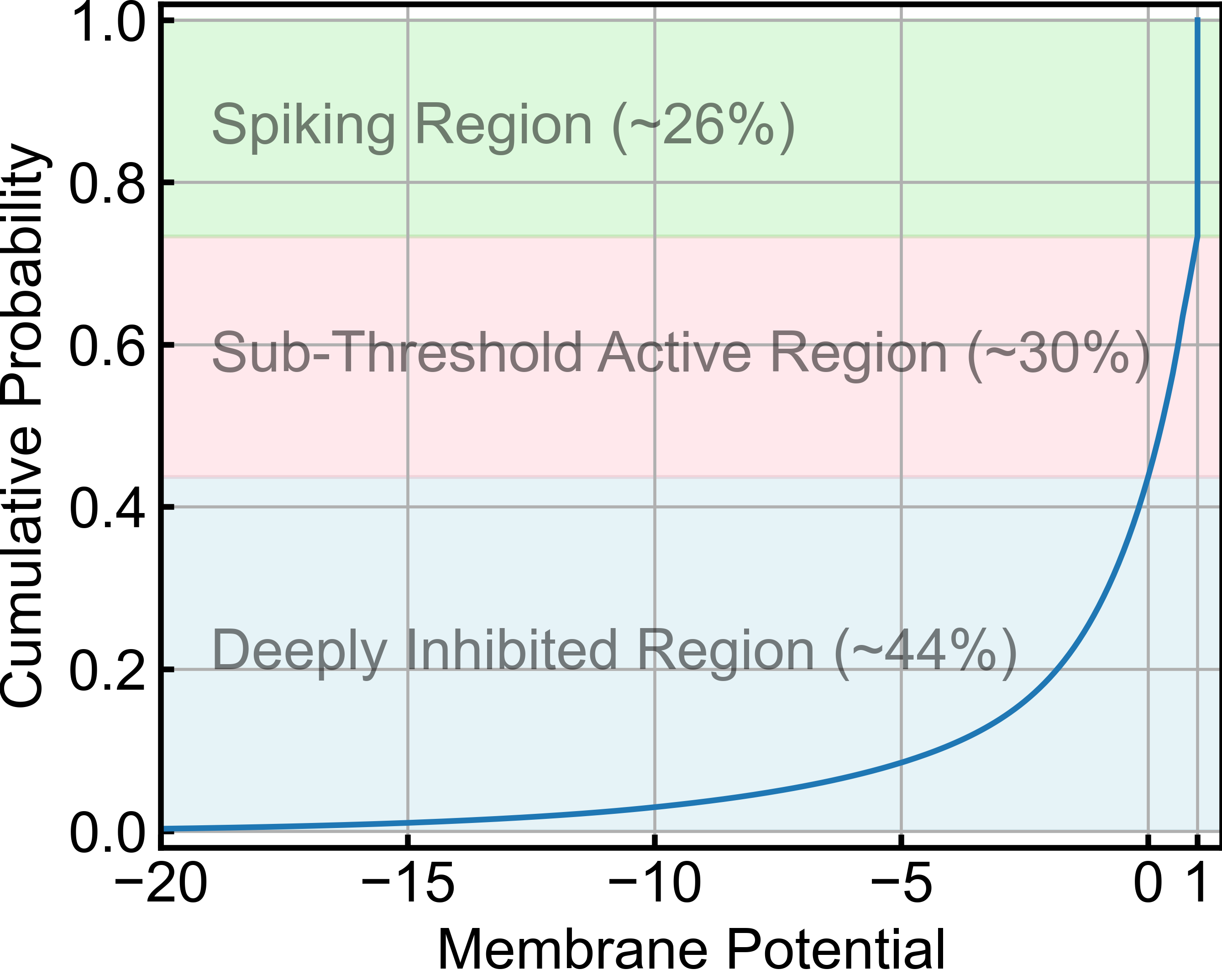}
    \caption{CDF of membrane potentials.}
    \label{fig:app_1}
  \end{subfigure}
  \centering
  \hfill
  \begin{subfigure}{0.31\linewidth}
    \centering
    \includegraphics[width=\textwidth]{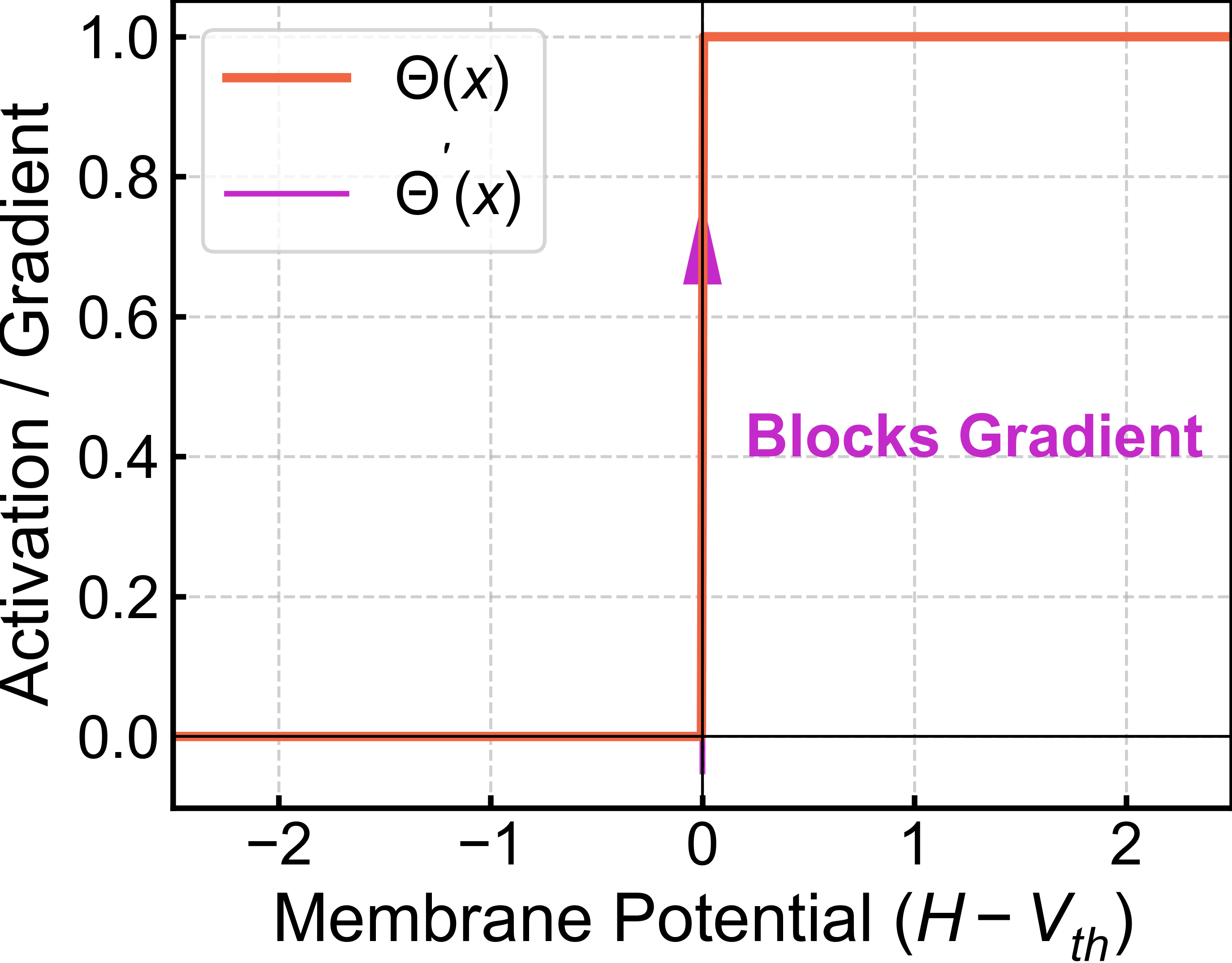}
    \caption{Heaviside step function.}
    \label{fig:app_2}
  \end{subfigure}
  \begin{subfigure}{0.31\linewidth}
    \centering
    \includegraphics[width=\columnwidth]{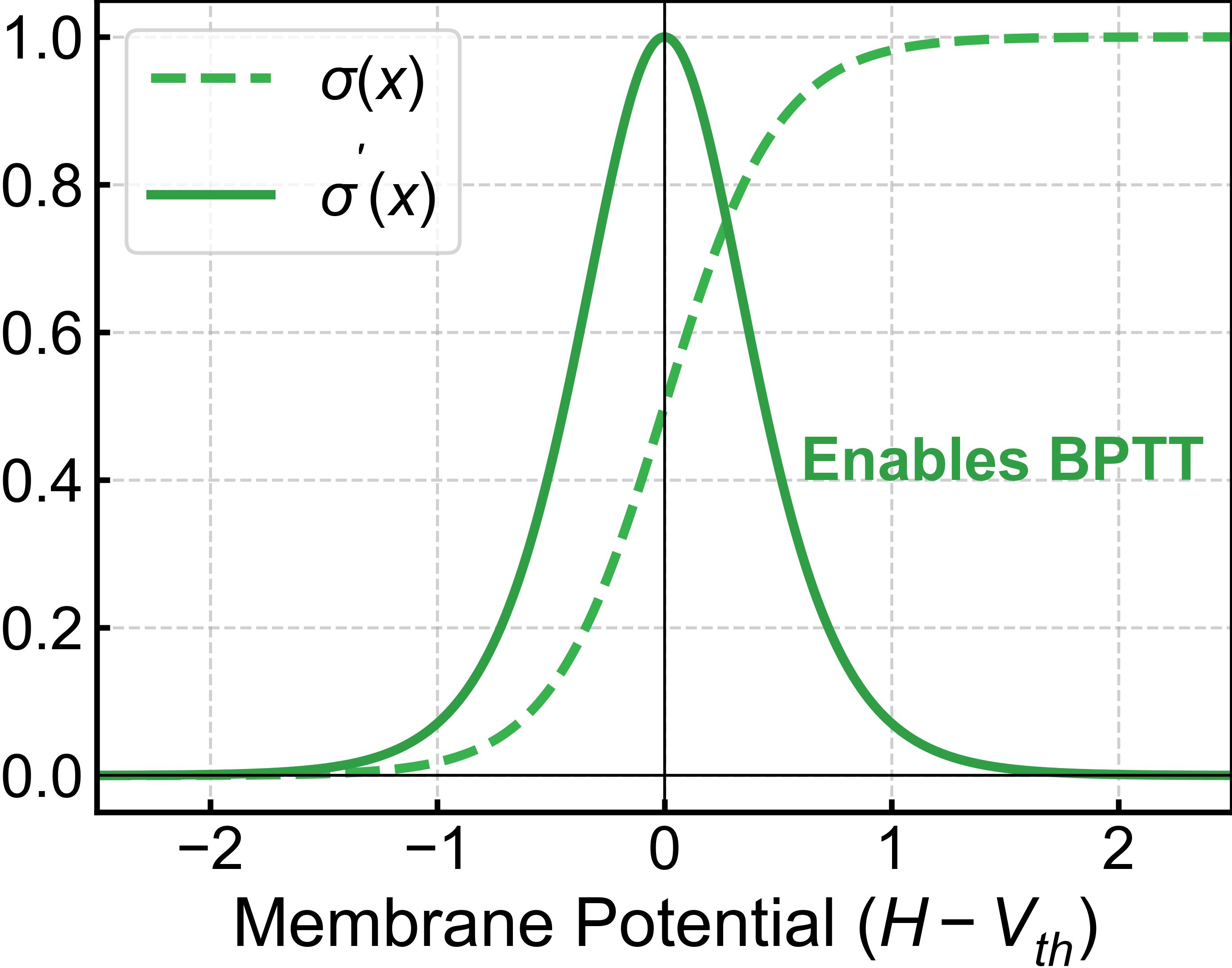}
    \caption{Surrogate function.}
    \label{fig:app_3}
  \end{subfigure}
  \caption{
  (a) The cumulative distribution function (CDF) of membrane potentials in a ResNet-18 SNN trained on ImageNet reveals a tripartite split. A substantial fraction (44\%) of neurons reside in the deeply inhibited region ($V_m \le 0$).
  (b) The Heaviside step function, whose derivative is zero almost everywhere, preventing gradient flow during backpropagation.
  (c) To enable gradient-based training, the true derivative is replaced by a smooth surrogate function $\sigma'(\cdot)$ during the backward~pass.}
  \label{fig:app1}
\end{figure}

\section{Supplementary Analysis and Experimental Details}
\label{app:exp_details}

\subsection{Supplementary Visualizations and Analysis}
\label{app_c1}

\textbf{Motivation: Membrane Potential Distribution.}
As posited in the Introduction, the computational inefficiency in standard SNNs primarily stems from the continuous subthreshold state updates of temporarily or permanently inhibited neurons.
To quantify this, we tracked the membrane potential distribution of a ResNet-18 SNN model trained on ImageNet.
As illustrated by the cumulative distribution function (CDF) in \cref{fig:app_1}, the internal states of the neurons exhibit a tripartite distribution.
While a portion of the neurons resides in the active spiking region or the critical subthreshold active region ($0 < V_m < V_{th}$), 44\% of the neuronal population is found in the deeply inhibited region ($V_m \le 0$).
To quantify the recovery probability, we identified neurons whose membrane potentials fell below $D_{th}$ and measured whether they emitted at least one spike during the subsequent timesteps.
Averaged across layers and timesteps, approximately 1\% of all neurons subsequently fired.
These deeply inhibited neurons therefore have a low probability of firing in the immediate future, yet they continue to consume computational resources.
This observation motivates the introduction of the dormancy mechanism to dynamically bypass these redundant instance-level computations.

\textbf{Threshold Sensitivity Across Datasets and Architectures.}
To further examine the sensitivity of the selected dormancy thresholds, we conducted threshold sweeps with BPTT across multiple datasets and architectures.
As reported in \cref{table:app_dth}, each entry gives the classification accuracy and estimated energy reduction.
For the neuromorphic datasets, $D_{th}=-0.5$ provides a favorable performance-efficiency balance.
For the static datasets, $D_{th}=-0.05$ remains near the low-energy plateau.
These results support the threshold settings used in the main experiments.

\begin{table}[!t]
  \caption{
    Threshold-sweep results across datasets and architectures for BPTT-trained DSN models.
    Each entry reports accuracy (\%) and estimated energy reduction.
  }
  \label{table:app_dth}
  \small
  \centering
  \setlength{\tabcolsep}{4pt}
  \begin{tabular}{ccccc}
    \toprule
    $D_{th}$ &
    \makecell{DVS-CIFAR10\\(SD-Transformer)} &
    \makecell{DVS-Gesture\\(SD-Transformer)} &
    \makecell{ImageNet\\(ResNet-18)} &
    \makecell{CIFAR100\\(SD-Transformer)} \\
    \midrule
    $-0.01$ & $75.9\%$ ($\downarrow 71\%$) & $94.44\%$ ($\downarrow 65\%$) & $64.20\%$ ($\downarrow 73\%$) & $79.22\%$ ($\downarrow 44\%$) \\
    $-0.05$ & $76.6\%$ ($\downarrow 70\%$) & $95.48\%$ ($\downarrow 63\%$) & $64.44\%$ ($\downarrow 73\%$) & $79.36\%$ ($\downarrow 43\%$) \\
    $-0.1$  & $77.2\%$ ($\downarrow 68\%$) & $96.18\%$ ($\downarrow 60\%$) & $64.34\%$ ($\downarrow 72\%$) & $79.40\%$ ($\downarrow 42\%$) \\
    $-0.5$  & $77.9\%$ ($\downarrow 52\%$) & $96.88\%$ ($\downarrow 42\%$) & $64.50\%$ ($\downarrow 54\%$) & $79.50\%$ ($\downarrow 32\%$) \\
    $-1$    & $78.0\%$ ($\downarrow 38\%$) & $96.88\%$ ($\downarrow 30\%$) & $64.66\%$ ($\downarrow 37\%$) & $79.55\%$ ($\downarrow 21\%$) \\
    \bottomrule
  \end{tabular}
\end{table}

\textbf{Neuron Model and Surrogate Gradient Mechanism.}
Throughout all experiments, the networks utilize the Leaky Integrate-and-Fire (LIF) neuron model.
The intrinsic neuronal parameters were uniformly set across all architectures to ensure stability: the membrane time constant was set to $\tau = 2.0$, the firing threshold to $V_{th} = 1.0$, and the hard reset potential to $V_{\text{reset}} = 0.0$.
To circumvent the inherent non-differentiability of the Heaviside step function $\Theta(\cdot)$, which appears in both the spike-firing mechanism and the proposed dormancy gate, we replace its derivative during the backward pass with the continuous surrogate-gradient approximation shown in \cref{fig:app_3}.
Specifically, we derive the surrogate function $\sigma'(x)$ from the sigmoid function:
\begin{equation}
\sigma_d(x) = \sigma_s(x) = \sigma(x) = \frac{1}{1 + e^{-\alpha x}}
\end{equation}
\begin{equation}
\sigma'_d(x) = \sigma'_s(x) = \sigma'(x) = \frac{\partial \sigma(x)}{\partial x} = \alpha \cdot \sigma(x) \cdot (1 - \sigma(x))
\end{equation}
For all BPTT training configurations, the sharpness parameter was empirically set to $\alpha = 4.0$. This smooth approximation provides stable gradient flow for learning robust dormancy policies.

\begin{figure}[!t]
  \begin{subfigure}{0.49\linewidth}
    \centering
    \includegraphics[width=0.8\textwidth]{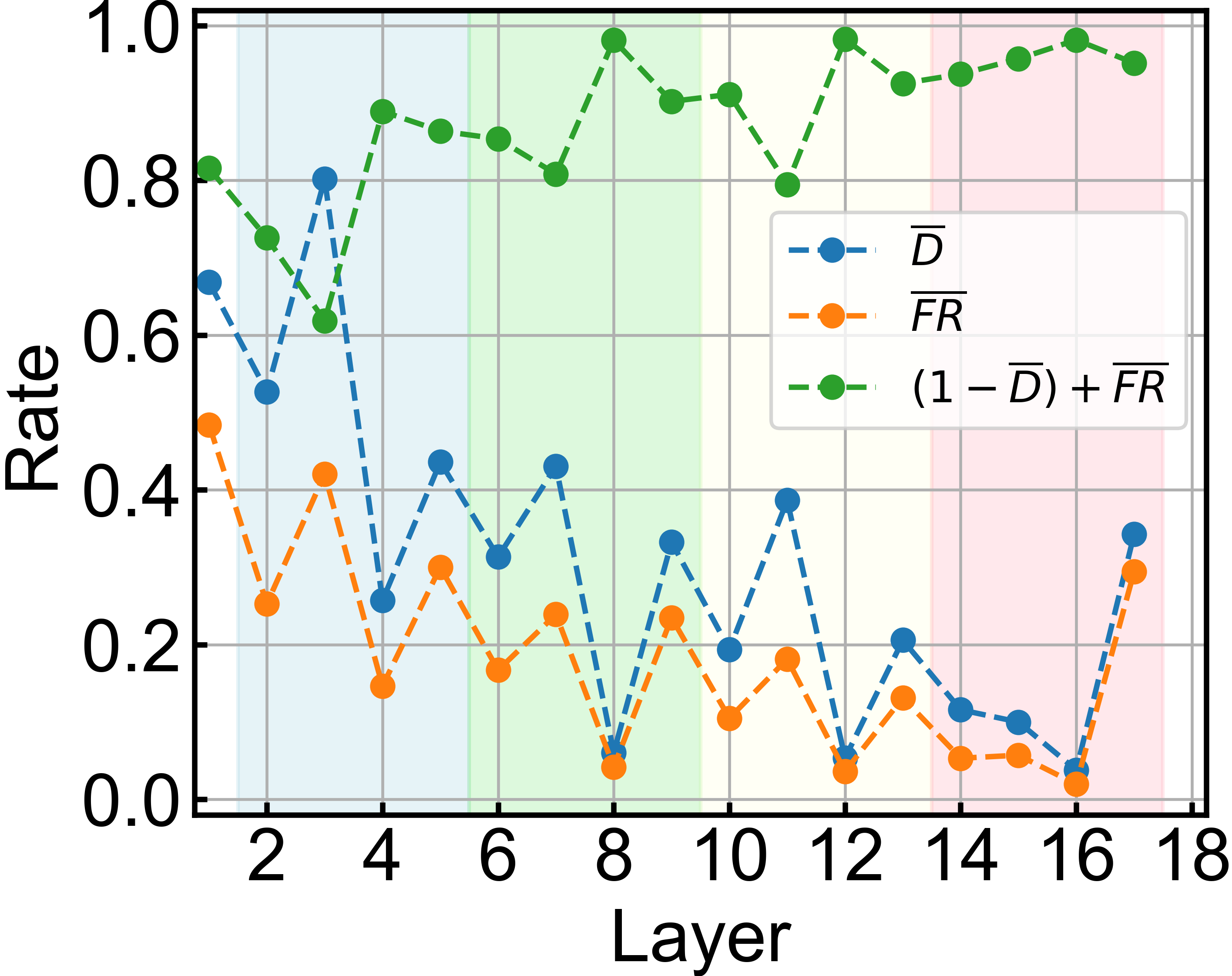}
    \caption{Firing rate, non-dormant rate, and NCU curves.}
    \label{fig:app_4}
  \end{subfigure}
  \centering
  \hfill
  \begin{subfigure}{0.49\linewidth}
    \centering
    \includegraphics[width=0.8\textwidth]{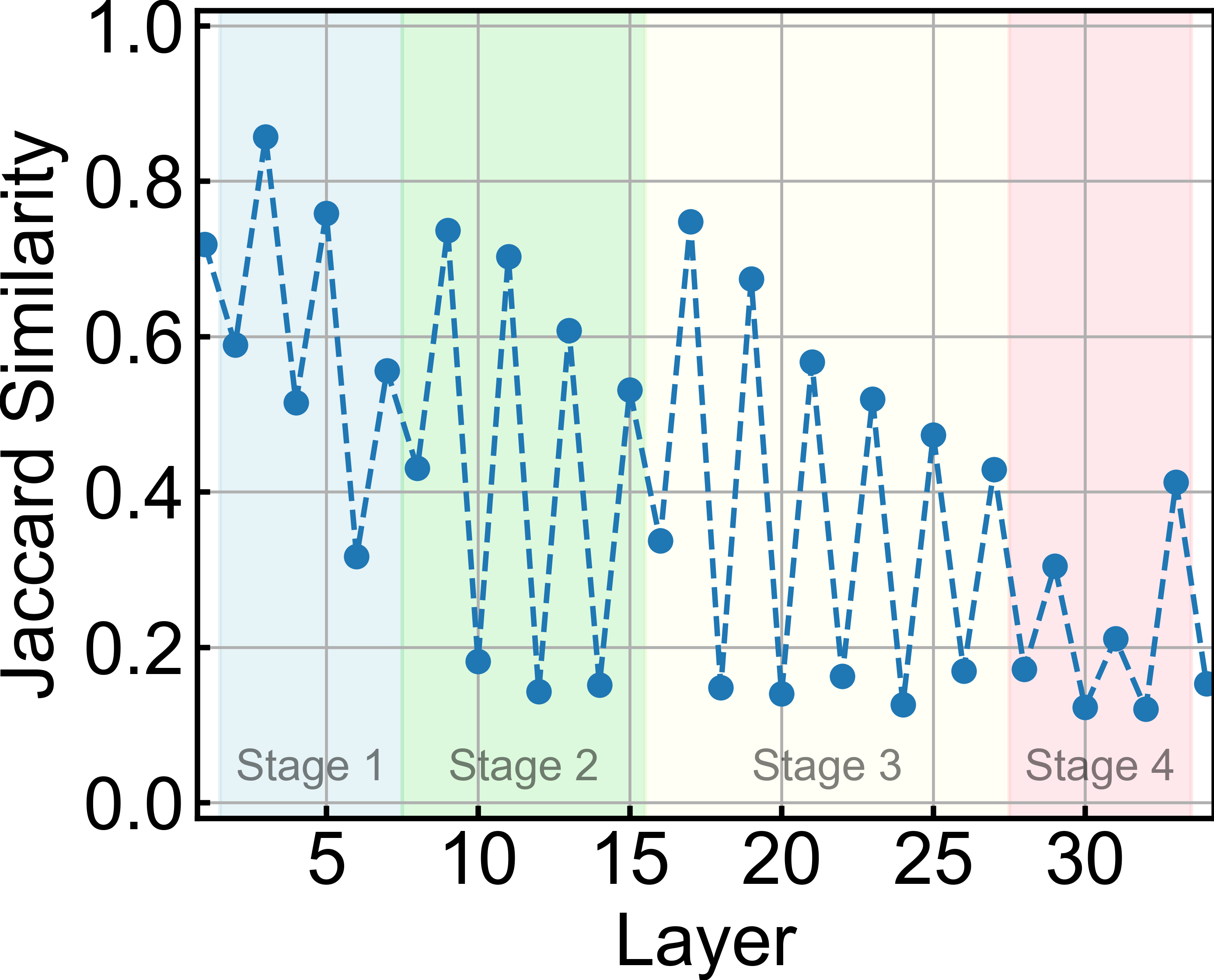}
    \caption{Layer-wise Jaccard similarity.}
    \label{fig:app_5}
  \end{subfigure}
  \caption{
  (a) Layer-wise firing rates, non-dormant rates, and the Neuronal Computational Utilization (NCU) on a ResNet-18 architecture.
  (b) Layer-wise Jaccard similarity of dormancy masks for a ResNet-34 model, showing their heterogeneity across inputs.}
  \label{fig:app2}
\end{figure}

\textbf{Deep Dive into DSN.}
To further elucidate the sparsity characteristics of the DSN, \cref{fig:app_4} details the layer-wise firing rates, non-dormant rates, and the resulting NCU across a ResNet-18 model.
The high NCU values across layers indicate that the DSN bypasses computations for a large portion of deeply inhibited neurons without suppressing critical active firing.
Furthermore, \cref{fig:app_5} visualizes the layer-wise Jaccard similarity of the dormancy masks across diverse inputs on a ResNet-34 model.
For conventional static pruning techniques, the Jaccard similarity is inherently fixed at 1 across all layers, as the pruned structure remains constant regardless of the input.
By contrast, the DSN exhibits substantial mask heterogeneity across layers, with more than 12 layers showing a mean overlap below 0.2.
This supports the conclusion that the DSN leverages dynamic, instance-level transient redundancy and adapts its dormancy masks across samples.
This capacity for sample-specific dynamic pruning is consistent with the observed advantage over static pruning methods.

\subsection{Implementation Details and Training Protocols}
\label{app_c2}

\textbf{Hyperparameters and Data Augmentation.}
Standard data augmentation pipelines were applied equivalently to all baselines and DSN models to ensure a fair comparison.
The comprehensive training hyperparameters, including learning rates, optimizers, and batch sizes, were closely aligned with those used in the original Spike-Driven Transformer to maintain consistency. A detailed summary of these key configurations is provided in \cref{table:hyperparams_supp}.

\begin{table}[!t]
    \small
    \caption{Training hyperparameters. A summary of the key hyperparameters utilized across all datasets and network architectures.}
    \label{table:hyperparams_supp}
    \renewcommand{\arraystretch}{0.8}
    \centering
    \begin{tabular}{lcccccc}
        \toprule[1.5pt]
        Dataset & Architecture & \makecell{Learning\\Rate} & \makecell{Weight\\Decay} & \makecell{Batch\\Size} & Epochs & Optimizer \\
        \midrule[1pt]
        \multirow{2}{*}{CIFAR} & MS-ResNet & 2e-2 & 5e-4 & 64 & 300+10 & SGD \\
        & SD-Transformer-2-512 & 3e-4 & 6e-2 & 64 & 300+10 & AdamW \\
        \cmidrule{2-7}
        \multirow{2}{*}{ImageNet} & MS-ResNet & 5e-2 & 2e-5 & 64 & 150+5 & SGD \\
         & SD-Transformer-8-384 & 7.5e-4 & 1e-2 & 28 & 300+10 & LAMB \\
        \cmidrule{2-7}
        DVS-Gesture & SD-Transformer-2-256 & 1e-4 & 6e-2 & 16 & 200+10 & LAMB \\
        \cmidrule{2-7}
        DVS-CIFAR10 & SD-Transformer-2-256 & 2e-3 & 1e-4 & 32 & 200+10 & AdamW \\
        \bottomrule[1.5pt]
    \end{tabular}
\end{table}

\textbf{Training Strategy and Implementation.}
Under the proposed BPTT paradigm, the DSN models were trained entirely from scratch on the static CIFAR10, CIFAR100, and large-scale ImageNet datasets, as well as the neuromorphic DVS-CIFAR10 dataset.
Due to the relatively small size of the DVS-Gesture dataset, the DSN model was fine-tuned from a pre-trained SNN baseline checkpoint to ensure stable convergence.
To optimize memory utilization and accelerate the training phase, automatic mixed precision was leveraged across all experiments.
For the main experiments and single-seed ablations, we used a fixed random seed of 2025.
Multi-seed analyses used the three random seeds described in the main text.

\textbf{Hardware, Software, and Baselines.}
All computational experiments were conducted on an internal Linux server running Ubuntu 22.04 with two NVIDIA RTX 5090 GPUs.
The software environment used Python 3.10, PyTorch 2.8.0, and SpikingJelly 0.0.0.0.15.
For CIFAR10, CIFAR100, DVS-Gesture, and DVS-CIFAR10, baseline results were reproduced internally by running the original, unmodified code under identical hardware and software conditions.
Because reproducing ImageNet baselines is computationally costly, the ImageNet baseline results were taken from the published literature for comparison.



\end{document}